\documentclass[11pt,letterpaper]{article}

\usepackage[margin=1in,top=1.1in,bottom=1in,headheight=28pt,headsep=14pt]{geometry}
\usepackage[T1]{fontenc}
\usepackage[utf8]{inputenc}
\usepackage{amsmath,amssymb}
\usepackage{newtxtext,newtxmath}
\usepackage{microtype}

\usepackage[table]{xcolor}
\definecolor{deepnavy}{HTML}{1F3B73}
\definecolor{warmorange}{HTML}{FF7B3A}
\definecolor{softgray}{HTML}{666666}
\definecolor{rulelinegray}{HTML}{B3B3B3}
\definecolor{abstractbg}{HTML}{F2F4F6}
\definecolor{paleblue}{HTML}{EDF3FC}
\definecolor{paleorange}{HTML}{FFF2EB}

\usepackage[explicit]{titlesec}
\titleformat{\section}
  {\fontsize{16}{20}\selectfont\bfseries\color{deepnavy}}{\thesection}{0.6em}{#1}
\titleformat{\subsection}
  {\fontsize{13}{16}\selectfont\bfseries\color{deepnavy}}{\thesubsection}{0.5em}{#1}
\titleformat{\subsubsection}
  {\normalsize\bfseries\color{deepnavy}}{\thesubsubsection}{0.5em}{#1}
\titlespacing*{\section}{0pt}{1.4em}{0.6em}
\titlespacing*{\subsection}{0pt}{1.1em}{0.45em}
\titlespacing*{\subsubsection}{0pt}{0.9em}{0.35em}
\titleformat{\paragraph}[runin]
  {\normalsize\bfseries\color{deepnavy}}{}{0pt}{#1.}
\titlespacing*{\paragraph}{0pt}{0.8em}{0.5em}

\usepackage{graphicx}
\usepackage{tikz}
\usepackage{adjustbox}
\usepackage{float}
\usepackage{caption}
\usepackage{booktabs}
\usepackage{array}
\usepackage{tabularx}
\usepackage{longtable}
\usepackage[most]{tcolorbox}

\usepackage{enumitem}
\setlist{nosep,leftmargin=1.5em,labelsep=0.6em}
\usepackage{placeins}
\usepackage{xurl}
\usepackage{fancyhdr}
\usepackage{lastpage}
\fancypagestyle{firstpage}{%
  \fancyhf{}
  \fancyhead[L]{\includegraphics[height=14pt]{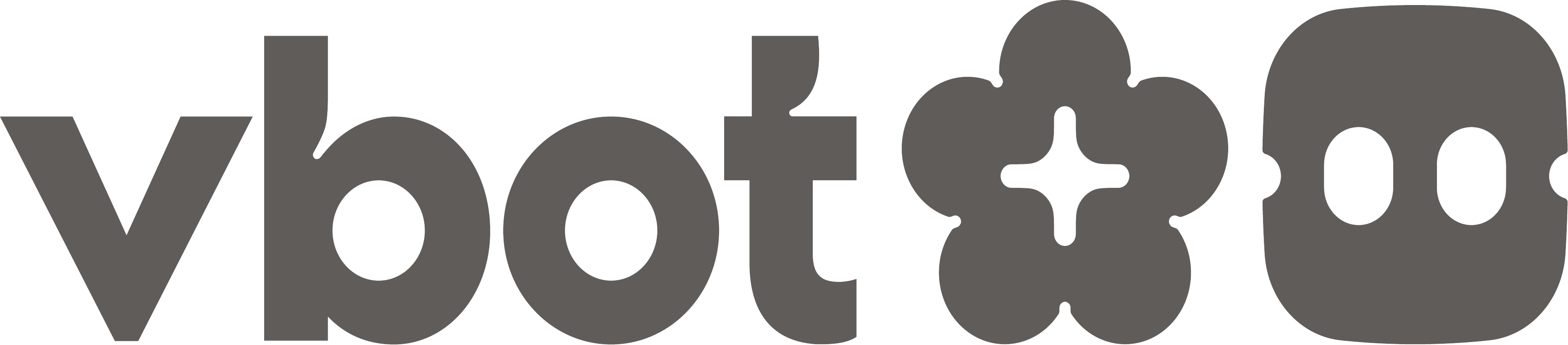}}
  \fancyhead[R]{\textcolor{softgray}{\footnotesize\itshape November 2026 \\ Vbot Physical AI Lab}}

  \fancyfoot[C]{\textcolor{softgray}{\footnotesize\thepage}}
}

\usepackage{hyperref}
\hypersetup{
  colorlinks=true,
  linkcolor=deepnavy,
  citecolor=deepnavy,
  urlcolor=deepnavy,
  pdftitle={VBot Exo2Ego Policy: Lightweight Data Collection and Exocentric-to-Egocentric Video Generation},
  pdfsubject={VBot Exo2Ego Policy technical report},
  pdfauthor={Tong Rao; Zhihao Zhang; Yuntian Wang; Siyu Zhou; Hailong Qin; Yinan Yu},
}

\newcommand{\reporttitleblock}{%
  \thispagestyle{firstpage}
  \vspace*{-0.4em}
  {\noindent\fontsize{25}{29}\selectfont\bfseries\color{deepnavy}
    VBot Exo2Ego Policy\par}
  \vspace{0.2em}
  {\noindent\fontsize{16}{20}\selectfont\bfseries\color{deepnavy}
    Lightweight Data Collection and\\
    Exocentric-to-Egocentric Video Generation\par}
  \vspace{0.45em}
  {\noindent\fontsize{11}{13}\selectfont\color{deepnavy}
    Tong Rao\textsuperscript{*}\quad
    Zhihao Zhang\textsuperscript{*}\quad
    Yuntian Wang\textsuperscript{*}\quad
    Siyu Zhou\textsuperscript{$\dagger$}\quad
    Hailong Qin\textsuperscript{$\dagger$}\quad
    Yinan Yu\textsuperscript{$\dagger$}\par}
  \vspace{0.15em}
  {\noindent\fontsize{10.5}{12.5}\selectfont
    \textbf{\color{deepnavy}Vbot Physical AI Lab}\quad
    \textcolor{softgray}{\textit{Shanghai, China}}\par}
  \vspace{0.1em}
  {\noindent\fontsize{8.5}{10}\selectfont
    \begin{tabularx}{\linewidth}{@{}XXX@{}}
      \href{mailto:tong.rao@vbot.cn}{tong.rao@vbot.cn} &
      \href{mailto:zhihao.zhang@vbot.cn}{zhihao.zhang@vbot.cn} &
      \href{mailto:26120249@bjtu.edu.cn}{26120249@bjtu.edu.cn} \\
      \href{mailto:siyu.zhou@vbot.cn}{siyu.zhou@vbot.cn} &
      \href{mailto:hailong.qin@vbot.cn}{hailong.qin@vbot.cn} &
      \href{mailto:yinan.yu@vbot.cn}{yinan.yu@vbot.cn}
    \end{tabularx}\par}
  {\noindent\fontsize{8.5}{10}\selectfont\color{softgray}
    \textsuperscript{*}Equal contribution.\quad
    \textsuperscript{$\dagger$}Corresponding authors.\par}
  \vspace{0.2em}
  {\color{rulelinegray}\hrule height 0.4pt}
  \vspace{0.4em}
}

\renewenvironment{abstract}{%
  \begin{tcolorbox}[
    colback=abstractbg,
    colframe=abstractbg,
    boxrule=0pt,
    arc=2mm,
    left=14pt,right=14pt,top=10pt,bottom=12pt,
    before skip=0.3em,after skip=0.8em
  ]
  {\centering\large\bfseries\color{deepnavy} Abstract\par}
  \vspace{0.5em}
  \setlength{\parskip}{0pt}
  \noindent\fontsize{10.5}{13.5}\selectfont
}{%
  \par\end{tcolorbox}
}
\begin{document}
\reporttitleblock

\begin{abstract}
We present VBot Exo2Ego Policy, a video generation model that turns third-person videos into first-person observations of a selected actor, offering a practical way to expand first-person operation data from abundant real-world footage. To support this goal, we introduce a lightweight data collection pipeline and mobile app that support first-person recording, third-person recording, and synchronized capture of both views. This setup combines accessible single-view collection with paired RGB and camera geometry for learning cross-view correspondence. Building on a pretrained video diffusion transformer, our model incorporates geometric guidance and auxiliary supervision from semantic masks and hand keypoints to facilitate optimization and improve the fidelity of hand–object interactions. Camera geometry is injected stochastically during training and gradually removed, allowing inference without target-camera calibration. To accommodate the diversity of Internet videos, we combine source-video perception and reasoning, optional actor references, and multi-shot adaptation to identify the intended actor and follow an operation across camera cuts. The collection system and generation model provide a foundation for expanding first-person operation data across tasks, scenes, and participants.
\end{abstract}

\begin{figure}[H]
\centering
\includegraphics[width=0.86\linewidth]{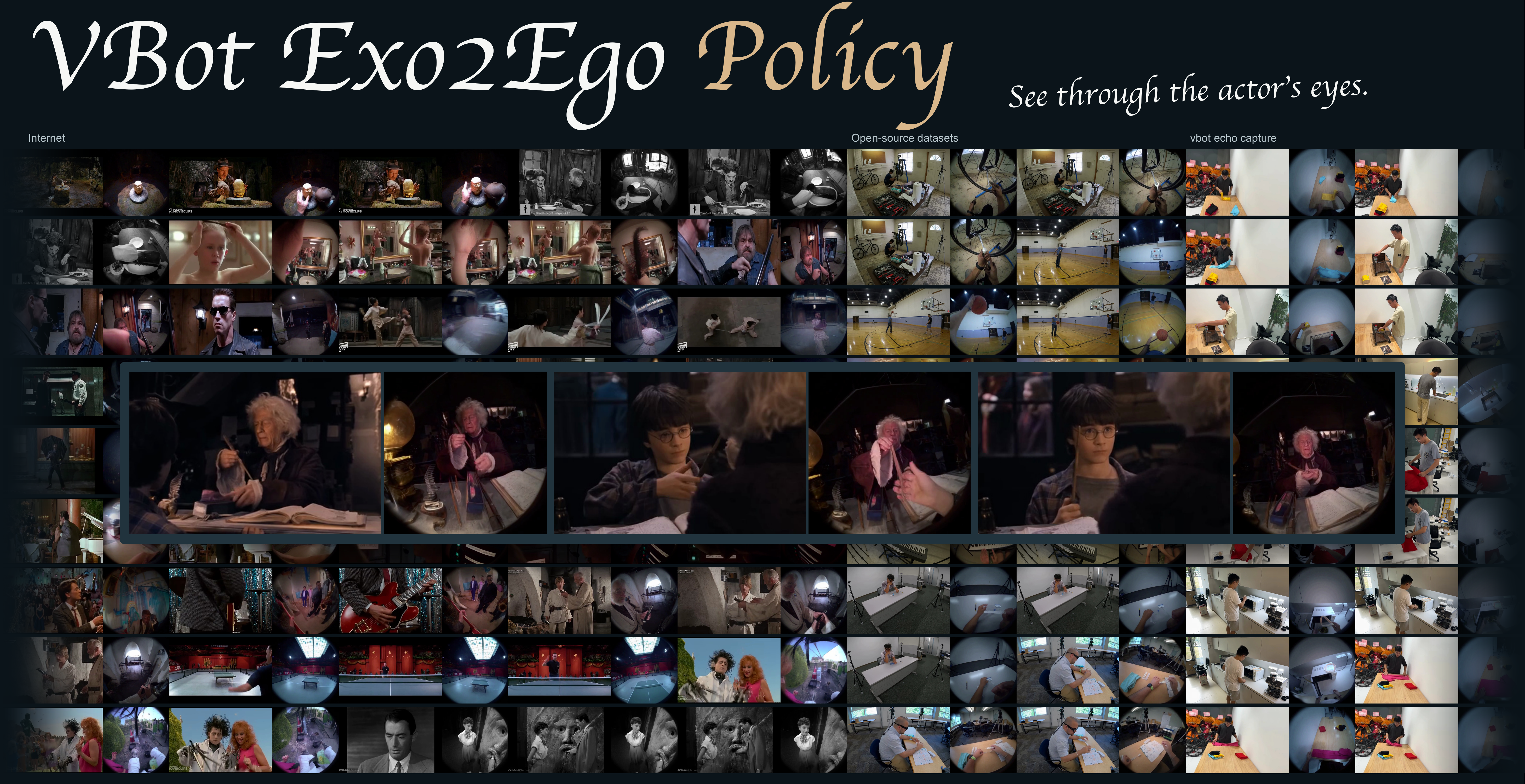}
\caption{\textbf{From observed operations to first-person video.} Selected EXO inputs and generated EGO frames.}
\label{fig:teaser}
\end{figure}

\clearpage

\section{Introduction}

\label{sec:introduction}

First-person video exposes the acting hands, manipulated objects, and
local workspace in a common visual context. It is useful for learning
operations, but collecting diverse first-person recordings is constrained
by capture equipment and participant effort. Third-person video offers
broader coverage, provided that view conversion preserves the actor,
interaction, and event sequence.

We address two connected problems. The first is a lightweight data
collection standard that accommodates both accurate paired supervision
and accessible single-view capture. Synchronized LiDAR-equipped iPhones
provide real EGO/EXO pairs, camera poses, and near-field geometry.
Crowdsourced participants can instead record either viewpoint with one
camera. Real EGO recordings support manipulation pretraining directly;
real EXO recordings provide inputs for generating EGO counterparts.

The second is an EXO-to-EGO model that preserves operation content across
viewpoints. VBot Exo2Ego Policy builds on a pretrained video diffusion
transformer and combines camera conditioning with auxiliary supervision
of hands and manipulated objects. Geometry guides camera-relative structure during training.
Semantic masks couple interaction regions with RGB generation, while
ACE features and 2D joint targets encourage hand structure that can be
recovered by downstream perception. Actor-reference images and
synchronized multi-view shot construction address multiple people and
camera changes in real videos.

Our contributions are:
\begin{enumerate}
\item \textbf{A lightweight collection and preprocessing standard.}
Precise paired capture and diverse single-view crowdsourcing share
view-aware processing, cross-view identity association, and data
selection suited to each supervision stage.
\item \textbf{A model for operation-preserving view conversion.}
Stochastic camera conditioning, semantic HOI supervision, and
ACE/GT hand supervision complement source perception, actor selection,
and multi-shot training. We evaluate RGB fidelity and the hand/object
observations extracted from generated videos.
\end{enumerate}

Section~\ref{sec:data-contract} presents collection and preprocessing.
Sections~\ref{sec:method-overview} and~\ref{sec:auxiliary-supervision}
describe the generator and interaction supervision.
Section~\ref{sec:evaluation} evaluates RGB fidelity and recoverable
hand--object structure.

\section{Related Work}
\label{sec:related-work}

\paragraph{Novel view synthesis and dynamic representations}
Novel view synthesis (NVS) has advanced through improvements in scene
representation, optimization efficiency, and generalizable reconstruction.
NeRF~\cite{mildenhall2020nerf} represents a static scene as a continuous
field of volume density and view-dependent radiance, learned from posed
images through differentiable volume rendering. NeuS~\cite{wang2021neus}
introduces a signed distance function and a volume-rendering formulation
that reduces geometric bias in surface reconstruction, improving the
recovery of scene surfaces. Instant-NGP~\cite{mueller2022instant} combines
multiresolution hash encoding, compact neural networks, and efficient GPU
kernels to substantially accelerate optimization and rendering.
3D Gaussian Splatting (3DGS)~\cite{kerbl2023gaussians} uses explicit
anisotropic Gaussians, adaptive density control, and differentiable
splatting to support high-quality, real-time novel-view rendering.

Recent feed-forward approaches predict renderable Gaussian representations
directly from images. Apple's SHARP~\cite{mescheder2025sharp} regresses a
metric 3D Gaussian representation from a single photograph in one network
pass, supporting real-time rendering of nearby viewpoints.
AnySplat~\cite{jiang2025anysplat} jointly predicts 3D Gaussians and camera
intrinsics and extrinsics from uncalibrated multi-view images, extending
feed-forward reconstruction to unconstrained image collections.

Operation videos additionally require representations of temporal change.
4D Gaussian Splatting (4D-GS)~\cite{wu2024gaussians4d} augments a shared
set of 3D Gaussians with a time-conditioned deformation model. A decomposed
4D neural voxel encoding captures spatial and temporal features, and a
lightweight decoder predicts Gaussian deformations at queried timestamps.
This formulation models scene structure and motion across time while
retaining efficient Gaussian rasterization, enabling real-time rendering
of dynamic scenes from novel viewpoints. For EXO-to-EGO conversion, these
advances motivate preserving scene motion together with camera-relative
geometry when changing viewpoints.

\paragraph{Feed-forward geometry and dynamic reconstruction}
DUSt3R~\cite{wang2024dust3r} learns pairwise point-map reconstruction;
VGGT~\cite{wang2025vggt} jointly predicts cameras, depth, point maps,
and tracks. Depth Anything 3 (DA3)~\cite{lin2025da3} unifies geometry prediction
through depth and camera rays, while $\pi^3$ (Pi3)~\cite{wang2026pi3} removes
the fixed reference-view choice with a permutation-equivariant
architecture. These models provide camera and scene estimates for
constructing view-aligned geometric conditions.
Recent dynamic models extend this foundation: D4RT~\cite{zhang2025d4rt}
jointly recovers geometry, cameras, and spatiotemporal correspondence;
4RC~\cite{luo2026fourrc} encodes a video once and queries geometry and
motion across timestamps; Any4D~\cite{karhade2025any4d} predicts dense
metric geometry and scene flow, optionally using additional sensors.
These advances improve the temporal geometric support available to
video generation. First-person synthesis additionally requires
appearance in newly exposed regions, fine hand articulation, and
hand--object contact that may be weakly observed in EXO footage.

\paragraph{Video reshooting and camera control}
Video reshooting preserves the content and dynamics of an input video
while changing its camera trajectory. Geometric proxies and learned
video conditions provide complementary support for generating the new
views.
TrajectoryCrafter~\cite{yu2025trajectorycrafter} combines point-cloud
renders and source video in a dual-stream diffusion model. For
world-consistent camera-controlled generation,
GEN3C~\cite{ren2025gen3c} renders a persistent 3D cache to guide newly
generated frames. EX-4D~\cite{hu2025ex4d} uses a depth watertight mesh
to address occlusion and boundary artifacts under extreme viewpoint
changes. A complementary approach,
ReCamMaster~\cite{bai2025recammaster}, learns source-video and camera
conditioning from synchronized synthetic multiview videos to reproduce
the input dynamics along a new camera trajectory.

Vista4D~\cite{lin2026vista4d} grounds
the source video and target cameras in a 4D point cloud, combining
persistent static content with reconstructed scene dynamics. Training
on reconstructed multiview dynamic videos improves robustness to
real-world depth and point-cloud artifacts while source-video
conditioning preserves appearance.
Manifold4D~\cite{mao2026manifold4d} further studies conflicts between
point-cloud renders and source-video conditions. It injects the
target-view render into the initial flow-matching noise, using it once
at initialization while retaining source video as the visual condition
during denoising. The render remains required to initialize inference.
Projected geometric conditions provide viewpoint cues but can contain
holes, occlusion errors, or distorted surfaces.
Reshooting therefore requires spatial alignment, temporal consistency,
and generative completion to preserve the observed event while filling
content newly exposed by camera changes.

\paragraph{Exocentric-to-egocentric video generation}
EgoX~\cite{kang2025egox} specializes video diffusion for single-video
EXO-to-EGO conversion through LoRA adaptation, rendered EGO priors,
and geometry-guided self-attention. The rendered prior provides target
viewpoint cues, while source-video conditioning carries appearance and
action evidence. Large EXO/EGO viewpoint changes can expose regions
missing from that prior, making completion of the acting hands and
local workspace particularly important.
Grounded-Exo2Ego~\cite{wang2026groundedexo2ego} combines geometric
anchoring with an object-level semantic grounding branch to restore
content under unreliable projections. Its camera re-localization
improves alignment between reconstructed geometry and paired EGO
targets, and its synthetic data engine supplies additional aligned
supervision. These works establish geometry, semantics, and training
alignment as complementary ingredients of first-person generation.
Our model learns camera guidance through stochastic ray conditioning
that is annealed away across training updates. Joint semantic-mask
prediction and ACE/GT hand supervision connect RGB generation with
interaction regions and extractable joint structure, while inference
uses EXO video, text, and an optional actor reference.

\paragraph{Paired data and interaction supervision}
Ego-Exo4D~\cite{grauman2024egoexo4d} supplies synchronized first- and
third-person observations. Our collection standard combines paired
phone RGB-D capture with single-view crowdsourcing. ACE-Ego-Hand
features~\cite{liu2026ace} and hand annotations from
HOT3D~\cite{banerjee2025hot3d}, ARCTIC~\cite{fan2023arctic}, and
Re:InterHand~\cite{moon2023reinterhand} support interaction-aware
supervision and hand readouts.

\section{Data Collection and Preprocessing}
\label{sec:data-contract}
\label{sec:system-overview}
\label{sec:data-preprocessing}

We develop a data collection pipeline and a mobile capture app supporting
EGO, EXO, and synchronized paired recording. The pipeline connects
reliable paired capture with scalable first-person data expansion:
real EGO/EXO pairs provide cross-view training targets and geometric
observations, real EGO recordings directly support manipulation
pretraining, and captured or Internet EXO videos provide inputs to
VBot Exo2Ego Policy for EGO generation. The app, capture protocol,
and data standard connect these sources to the overall model workflow.
Table~\ref{tab:acquisition-routes} summarizes their inputs and uses;
recorded and generated views retain distinct provenance.

\subsection{Capture App and Acquisition Routes}
\label{sec:capture-app}

\begin{table}[H]
\centering
\caption{\textbf{Data sources and uses.} Recorded and generated
views have distinct provenance.}
\label{tab:acquisition-routes}
\small
\renewcommand{\arraystretch}{1.16}
\begin{tabularx}{\linewidth}{@{}p{0.20\linewidth}p{0.31\linewidth}X@{}}
\toprule
\textbf{Route} & \textbf{Recorded input} & \textbf{Use / derived product} \\
\midrule
Precise paired capture
& EGO and EXO RGB from paired LiDAR-equipped iPhone Pro devices; depth, pose, and timing metadata
& Real paired supervision for EXO-to-EGO training and held-out evaluation. \\
EGO capture
& A real first-person video
& Real EGO data for manipulation pretraining; a paired EXO view is not required. \\
EXO capture / Internet
& A real third-person video from contributors or the Internet
& Input to EXO-to-EGO generation; a real EXO / generated EGO pair after QC. \\
\bottomrule
\end{tabularx}
\end{table}

\paragraph{Synchronized paired RGB-D capture}
Two LiDAR-equipped iPhone Pro devices record the same operation from
EGO and EXO viewpoints, associating the same actor, manipulated objects,
and continuous action timeline. Temporal synchronization pairs frames
at the same point in the operation. ARKit supplies camera trajectories,
while LiDAR supplies near-field depth observations; the geometric
pairing procedure in Section~\ref{sec:geometric-pairing} relates the two
views in a common metric coordinate frame.

The app provides the recording interface, and a standard operating
procedure (SOP) specifies camera placement, mounting, imaging settings,
and visibility of the hands, objects, and key events. The EGO mount
permits natural head motion; after the common-start setup described
below, the EXO phone remains fixed during the operation. Before capture,
the operator rehearses the action and checks that its key stages remain
recognizable at the model's input resolution. Each recording retains
RGB, timestamps, camera intrinsics, ARKit trajectories and tracking
states, and available depth with registration, validity, and confidence
information. Frame selection, cropping, and resizing preserve mappings
to the source frames and the corresponding camera parameters.

\paragraph{Single-view collection and Internet EXO ingestion}
Participants can contribute EGO or EXO video to broaden task, object,
scene, and viewpoint coverage. The same ingestion route accommodates
large-scale Internet EXO footage, expanding the source pool beyond
newly commissioned capture. After manipulation-clip selection and
target-actor association, EXO footage is supplied to the generator to
produce the corresponding first-person view. This reduces dedicated
EGO capture effort when expanding manipulation data, without requiring
depth or camera calibration for each EXO input. Actor references and
multi-shot adaptation address multi-person scenes and camera cuts
(Section~\ref{sec:actor-reference-multishot}). Quality selection and generated
data review are described in the following subsections.

\subsection{Geometric Pairing and Cross-View Correspondence}
\label{sec:geometric-pairing}

\paragraph{Wide-baseline estimation}
During manipulation, EGO and EXO cameras often have a wide spatial
baseline, limited visual overlap, hand--object occlusions, and substantial
differences in scale and appearance. Relevant estimation approaches
include feed-forward geometry models such as VGGT~\cite{wang2025vggt},
Pi3~\cite{wang2026pi3}, DA3~\cite{lin2025da3}, and
DUSt3R~\cite{wang2024dust3r}, as well as learned matching and
reconstruction methods such as MASt3R~\cite{leroy2024mast3r},
SuperPoint+LightGlue~\cite{lindenberger2023lightglue},
LoFTR~\cite{sun2021loftr}, and RoMa~\cite{edstedt2024roma}.
In our low-overlap capture setting, direct recovery from widely
separated operational views can be unstable or fail, motivating a
capture protocol that supplies a visual connection between the streams.

\paragraph{Common-start SOP and joint reconstruction}
At the start of synchronized capture, the two cameras record a shared
visible region from nearby scene locations before moving to their
operational viewpoints. We retain these common-start observations and
jointly reconstruct them with keyframes from both EGO and EXO streams.
This creates a visual connection between sequences whose operational
views may have little overlap. In our experience, Pi3 and DA3 work well
under this capture arrangement. The common start supports cross-sequence
association; joint reconstruction and a metric reference then establish
the spatial pairing. The devices' independent ARKit coordinate systems
are not assumed to coincide.

\paragraph{ARKit-referenced validation and metric alignment}
We associate the jointly predicted EGO camera trajectory with the ARKit
EGO trajectory at matching timestamps and estimate a single
$\mathrm{Sim}(3)$ alignment for scale, rotation, and translation.
The aligned trajectory shape and residuals provide a consistency check
on predicted motion. For accepted reconstructions, the same global
transform is applied to both camera trajectories, and its scale factor
is applied consistently to predicted depths. For a predicted camera
center $\hat{c}$ and depth $\hat{D}$, this gives
$c=sR\hat{c}+t$ and $D^{\mathrm{metric}}=s\hat{D}$.
Relative EGO--EXO poses follow from the two camera poses in the shared
metric frame. ARKit trajectory agreement is considered alongside
geometric consistency in jointly observed regions; a successful
similarity fit alone does not establish the correctness of the EXO pose.
Temporal pairing, metric depth, and the shared camera frame together
support frame-level geometric correspondence in mutually visible regions.

\paragraph{LiDAR depth validation and preprocessing}
iPhone LiDAR depth also provides a near-field metric reference for
checking predicted depth and scale. ARKit scene depth is a processed,
low-resolution estimate with spatially varying reliability. In our
capture practice, reliability decreases substantially near or beyond
approximately five meters, so those measurements are not used as primary
validation anchors. This operating boundary varies with the device,
materials, lighting, and scene. Low resolution, finite projected spot
and sampling footprints, and interpolation can produce mixed
foreground/background estimates and merged fine structures at distant
objects or depth boundaries. Before validation, preprocessing aligns
RGB and depth timestamps, intrinsics, and image transforms, excludes
invalid, low-confidence, and out-of-range samples, and handles depth
discontinuities and outliers. Validation primarily uses valid near-field
observations while retaining their confidence and validity information.

After temporal and geometric alignment, we select video clips and
auxiliary targets according to their visual quality and intended training
role.

\subsection{Quality Assessment and Training-Stage Selection}

\label{sec:quality-selection}
The VLM assigns sampled frames clarity scores from 1 to 5, ranging from
severe degradation to sharp detail. Newly scored 141-frame clips use
eight uniformly spaced samples, including the endpoints. The clip
rating is the rounded arithmetic mean:
\[
\bar{s}=\frac{1}{N}\sum_{i=1}^{N}s_i,\qquad
s_{\mathrm{clip}}=\left\lfloor\bar{s}+\frac12\right\rfloor.
\]
The mean, minimum, and low-score count remain available for selection.
Normal fisheye distortion and optical black corners are part of the
camera model. View review additionally records wearer identity, EXO
head/orientation cues, and incidental EGO occlusion. Task hands and
objects entering the near field are treated as interaction content.

Low light and rapid motion reduce visible detail but contribute to
real-world diversity. We therefore select data according to its training
role (Table~\ref{tab:stage-quality-selection}). RGB supervision uses broad
coverage after anomaly rejection. Hand and semantic supervision exclude
insufficiently lit or excessively fast-motion samples to improve target
reliability.
\begin{table}[H]
\centering
\caption{\textbf{Data selection by supervision stage.}}
\label{tab:stage-quality-selection}
\small
\renewcommand{\arraystretch}{1.12}
\begin{tabularx}{\linewidth}{@{}p{0.19\linewidth}XX@{}}
\toprule
\textbf{Supervision} & \textbf{Selection} & \textbf{Purpose} \\
\midrule
RGB & Broad data coverage after hard anomaly rejection.
& Retain real-world lighting, motion, and appearance diversity. \\
Hand & Exclude insufficient illumination or excessively rapid motion.
& Improve feature and keypoint target reliability. \\
Semantic & Exclude insufficient illumination or excessively rapid motion.
& Improve region and boundary target reliability. \\
\bottomrule
\end{tabularx}
\end{table}

Hard-anomaly review covers the complete training clip. A single frame
with mosaics, anomalous blurring of essential content, a blackout, or
corrupt imagery rejects the sample; neither an average clarity score
nor trimming overrides that decision. Sampled clarity assessment and
complete-clip anomaly review have separate roles.

\paragraph{Task localization and supervision validation}
\label{sec:temporal-localization}
Video semantics, hand/object observations, motion, and masks localize the
continuous operation interval. Manipulation phases follow approach,
execution, and release, with task boundaries linked to the source frames.
Derived hand targets and masks are checked for valid observations,
missing or incorrect regions, boundary quality, and temporal consistency.
Each sample records RGB, geometry, hand, and semantic availability so
that usable RGB can contribute even when an auxiliary target is unreliable.

\subsection{First-Person Data Expansion and Provenance}
\label{sec:ingestion}

Recorded EGO directly supports manipulation pretraining. For EXO-only
sources, the trained generator produces an EGO counterpart using source
perception, text, and an optional actor reference. Review checks action
and object consistency, event order, and temporal coherence.

Each item retains its source interval, recorded/generated view identity,
model, prompt, and generation settings. Source-aware splits group all
derivatives of a recording together. Ingestion and any required privacy
processing preserve the actor and interaction evidence needed by the
intended training stage. Alignment and packaging are summarized in
Appendix~\ref{app:alignment-packaging}.

\section{VBot Exo2Ego Policy}

\label{sec:method-overview}

The VBot Exo2Ego Policy generates a first-person video of an operation from an exocentric (EXO) recording. The task presents two challenges: EXO footage varies in viewpoint, may contain multiple people, and can switch cameras mid-action, while the generated egocentric (EGO) video must faithfully depict the operation from the intended actor’s viewpoint. To handle this diversity, we use source-video reasoning and an optional actor reference image to identify whose view to generate, and train on clips that switch between synchronized EXO views so the model can follow the same action across cuts. To improve EGO fidelity, we use camera conditioning to guide learning of the target viewpoint and auxiliary semantic and hand supervision to preserve how the actor’s hands interact with objects.

\subsection{Source Conditioning and RGB Generation}
\label{sec:source-conditioning}

We use Qwen3.8-VL to describe the operation shown in an EXO clip \(X\).
Qwen receives frames \(F_X^{\mathrm{Qwen}}\) from the clip and a structured
perception record \(S_X\) containing actor identities, hand and object
regions, and interaction events. It produces a text description \(c\)
of the intended actor's actions and their temporal order.
\begin{equation}
    c = \operatorname{Qwen3.8\text{-}VL}
        \bigl(F_X^{\mathrm{Qwen}}, S_X\bigr).
    \label{eq:qwen-text}
\end{equation}

We adapt MiniMax-H3 Ref2VA for EXO-to-EGO video generation by training
an RGB LoRA while keeping the base model frozen. Let \(G_\theta\) denote
the adapted generator, with \(\theta\) denoting the RGB LoRA parameters.
The generator takes the EXO clip \(X\), the Qwen-generated description
\(c\) from Equation~\eqref{eq:qwen-text}, and an optional EXO screenshot
\(I_a\) identifying the actor whose first-person view is to be generated.
We write \(I_a = \varnothing\) when no screenshot is provided. Starting
from initial latent noise \(\epsilon_{\mathrm{init}}\), the generator
produces the EGO RGB video \(\hat{Y}\):
\begin{equation}
    \hat{Y} = G_\theta(X, c, I_a; \epsilon_{\mathrm{init}}).
    \label{eq:h3-generation}
\end{equation}

Within H3, we use the frozen video VAE to encode \(X\) as a visual
reference. The frozen H3 context encoder embeds source frames
\(F_X^{\mathrm{H3}}\) together with \(c\) and \(I_a\), when the reference
image is provided. We denote the resulting source conditions by
\(\mathcal{C} =
\bigl(E_{\mathrm{VAE}}(X),
      E_{\mathrm{ctx}}(F_X^{\mathrm{H3}}, c, I_a)\bigr)\).

For RGB training, let \(Y\) be the paired EGO target and
\(z_0 = E_{\mathrm{VAE}}(Y)\) its clean latent. At timestep \(t\), we
construct a noisy latent \(z_t\) by interpolating between \(z_0\) and
training noise \(\epsilon\) at noise level \(\sigma_t\). The H3 denoiser
takes \(z_t\), \(t\), and \(\mathcal{C}\) as inputs and predicts the
velocity \(\bar{v}_\theta(z_t,t;\mathcal{C})\). We train this prediction
with the weighted flow-matching objective
\begin{equation}
    \begin{aligned}
        z_t
        &= (1-\sigma_t)z_0 + \sigma_t\epsilon, \\
        \mathcal{L}_{\mathrm{RGB}}
        &= \mathbb{E}_{t,\epsilon}\!\left[
            w(t)
            \left\lVert
                \bar{v}_\theta(z_t,t;\mathcal{C})
                - (\epsilon-z_0)
            \right\rVert_2^2
        \right],
    \end{aligned}
    \label{eq:rgb-flow-matching}
\end{equation}
where \(w(t)\) is the scheduler weight and \(\bar{v}_\theta\) is the
effective velocity prediction. We freeze the base H3 model and train
the RGB LoRA to learn the EXO-to-EGO mapping.

\begin{figure}[H]
\centering
\includegraphics[width=\linewidth]{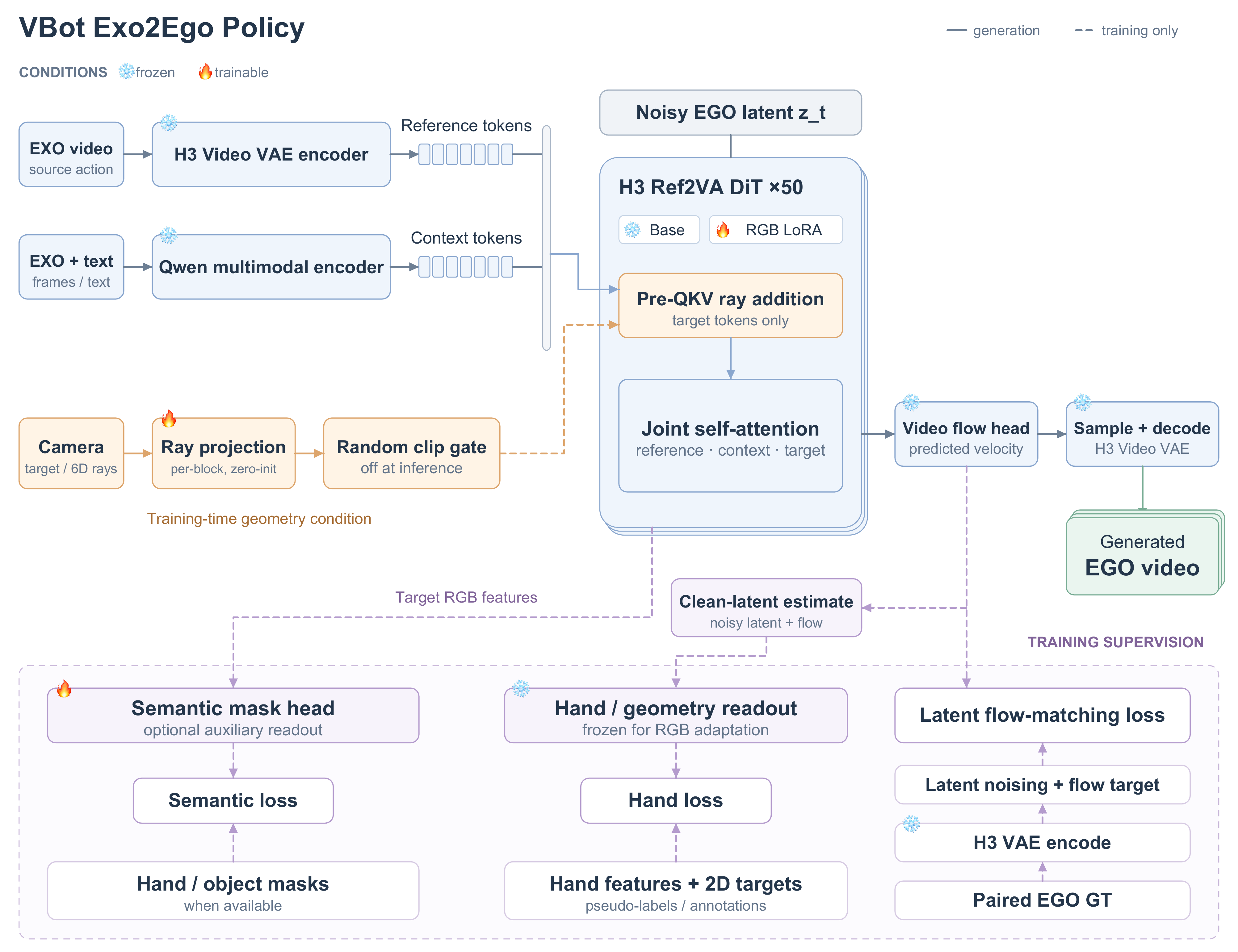}
\caption{\textbf{VBot Exo2Ego Policy architecture.} Solid paths generate EGO RGB; dashed paths introduce geometric, semantic, and hand supervision during training.}
\label{fig:model-architecture}
\label{fig:method-overview}
\end{figure}

\paragraph{Architecture}
Figure~\ref{fig:model-architecture} shows reference and context encoding,
RGB generation, and auxiliary supervision. Rays enter target tokens
during training. The semantic branch reads RGB features and returns
semantic gradients to the generator; hand supervision acts on clean
latent predictions or intermediate EGO features.

\subsection{Stochastic Camera Conditioning}

\label{sec:key-supervision}
\label{sec:geometry-conditioning}

To incorporate calibrated camera geometry into viewpoint learning
while preserving the pretrained generator's behavior at initialization,
we introduce stochastic camera conditioning. We sample a Bernoulli gate
for each training clip to control whether calibrated target-camera rays
are injected through zero-initialized projections. We anneal the
probability of ray injection to zero before the end of training, so
the final training phase uses the same conditioning inputs as inference:
the EXO clip, its text description, and an optional actor reference image.

\paragraph{Calibrated rays and token injection}
We represent the target camera geometry with Pl\"ucker rays. For a
calibrated fisheye camera, we use the camera model to compute a viewing
ray for each target EGO token. Let \(o\) be the camera center and \(d\)
the unit viewing direction, both expressed in the selected EXO
coordinate frame. We write each ray as
\begin{equation}
    r = [o \times d,\, d].
    \label{eq:plucker-ray}
\end{equation}
We collect these rays in \(R\) in the same order as the EGO tokens.

During training, each DiT block \(\ell\) uses a zero-initialized linear
projection \(W_\ell\) to map the rays to token embeddings. Before the
QKV projection, we add each ray embedding to its corresponding EGO
token, controlled by a binary gate \(b\).
\begin{equation}
    {u'}_{\ell,\mathrm{EGO}}
    = u_{\ell,\mathrm{EGO}} + b\,W_\ell R,
    \qquad b \in \{0,1\}.
    \label{eq:ray-injection}
\end{equation}
Here, \(u_{\ell,\mathrm{EGO}}\) and \({u'}_{\ell,\mathrm{EGO}}\) are
the EGO tokens before and after ray injection. We sample one gate
\(b\) per clip and share it across all frames, token positions, and
DiT blocks.

\paragraph{Condition dropout and annealing}
The ray probability follows training progress $q$:
\begin{equation}
b\sim\mathrm{Bernoulli}(p(q)),\qquad
p(q)=\begin{cases}
0.5,&0\leq q<0.7,\\
0.5\,\dfrac{0.8-q}{0.1},&0.7\leq q<0.8,\\
0,&0.8\leq q\leq1.
\end{cases}
\label{eq:ray-keep-schedule}
\end{equation}
The first 70\% of training exposes the model to both conditioned and
unconditioned clips. Probability decreases linearly over 70--80\%,
and the final 20\% is ray-free. Subsequent ray-free continuation retains
the learned RGB adaptation. Deployment bypasses the ray branch.

\subsection{Source Perception and Reasoning}
\label{sec:source-perception}

An EXO clip can show several people interacting with different objects.
To specify which operation the EGO video should depict, we construct
a text condition \(c\) that describes the intended actor's hand actions,
the objects involved, and the order of events. We use the same procedure
to prepare this text for training and inference.

\paragraph{Structured hand/object evidence}
For an EXO clip \(X\), we organize the source observations into a
structured record \(S_X\). The record identifies actors and objects
and stores hand roles, keyframe indices, two-dimensional regions,
and hand--object relations. Person tracks link observations across
frames to the same actor. We provide \(S_X\) and the EXO frames
\(F_X^{\mathrm{Qwen}}\) to Qwen3.8-VL as in
Equation~\eqref{eq:qwen-text}.

\paragraph{Reasoning text and keyframe ROIs}
In the detailed text format, we include keyframe evidence alongside
the natural-language description. The regions of interest (ROIs) are
bounding boxes in normalized EXO image coordinates. The interaction
relation identifies the actor, the hand, the action, and the object.
The example below describes a right-hand grasp at frame 70.

\begin{tcolorbox}
[colback=paleblue,colframe=deepnavy,boxrule=0.6pt,arc=1mm]
\small\raggedright
\textbf{EXO frame 70}\\
\texttt{actor=2; right\_hand.box=(0.42,0.38,0.63,0.72)}\\
\texttt{cup.box=(0.52,0.40,0.68,0.63)}\\
\texttt{relation=(actor\_2,right\_hand,grasps,cup)}\\
\textbf{Description.} The actor in the blue shirt grasps the cup
with the right hand, lifts it, and places it on the table.
\end{tcolorbox}

Qwen3.8-VL uses the source frames and the structured evidence to
generate the description. The H3 context encoder then embeds the
resulting text \(c\) together with the source visual inputs, as
described in Section~\ref{sec:source-conditioning}.

\paragraph{Concise task text}
The same operation can also be expressed as a short task description,
such as ``Grasp the cup, lift it, and place it on the table.'' Both
formats preserve the task and the event order. The detailed format
also specifies where the hands and objects appear at keyframes and
how they interact. The source associations and, when provided, the
actor reference image identify whose first-person view H3 should
generate.

\subsection{Actor Reference and Multi-Shot Adaptation}
\label{sec:actor-reference-multishot}

Internet EXO footage can show several people interacting in the same
workspace and can switch cameras while an operation is still in
progress. In a multi-person scene, the same clip can support several
first-person viewpoints. At a camera cut, the framing and viewing angle
can change abruptly even though the underlying action continues. We
use an optional actor reference to specify whose view to generate and
train on synchronized camera switches to maintain continuity across
source shots.

\paragraph{Actor reference}
We use an EXO screenshot as the optional reference image \(I_a\) in
Equation~\eqref{eq:h3-generation}. The screenshot can retain the wider
scene or be cropped around the intended actor. The reference image
identifies the person whose viewpoint the generator should follow,
while the EXO clip supplies the motion and timing of the operation.

During training, the reference image depicts the wearer of the paired
EGO camera. We align the reference image, the source perception record
\(S_X\), and the text description \(c\) with that same person. This
association links the visual reference to the hands, objects, and
actions described in the text.

\paragraph{Synchronized multi-view shot construction}
We construct multi-shot training inputs from EXO cameras that record
the same operation in sync. At selected points in the source timeline,
we switch from one EXO camera to another while keeping a continuous
EGO target from the same wearer. Each source frame therefore remains
aligned in time with its corresponding EGO frame across the cut.
These pairs teach the generator to follow a single operation as the
source viewpoint changes.

\paragraph{Inference}
We provide the EXO clip and its source-grounded text to the adapted
generator. In scenes with multiple people, the user can also provide
a screenshot to specify the intended actor. The generator produces
a continuous EGO RGB sequence for that actor across the source shots.
The joint RGB--semantic model can also output the corresponding
semantic masks.

\section{Auxiliary Supervision for Hand--Object Interaction}
\label{sec:auxiliary-supervision}

A generated EGO video can match the overall appearance of an operation
while misplacing a hand or object or distorting the motion of individual
joints. To improve hand and object localization, we supplement the RGB
objective with semantic masks that specify the regions occupied by the
wearer's hands and the manipulated objects. To preserve hand articulation
and motion, we use ACE features and annotated 2D keypoints as additional
supervision. Both objectives shape the RGB representation, and we
evaluate their effects on RGB fidelity and on the hand and object
observations recovered from the generated videos.

\subsection{Joint RGB and Semantic-Mask Supervision for HOI}
\label{sec:semantic-method}

We use semantic masks that are spatially and temporally aligned with
the EGO RGB video. Each mask labels pixels as background, manipulated
objects, or the camera wearer's hands, arms, and sleeves. The RGB and
mask branches share a frozen backbone, with a separate LoRA adapter
for each branch. The mask branch uses features from the RGB branch
to localize the labeled regions and track their motion over time.
During joint training, gradients from the mask loss flow through
these features to the RGB LoRA, so region-level supervision directly
shapes the representation used for RGB generation.

\paragraph{Mask supervision.}
Let \(M\) be the ground-truth semantic-mask video. We encode it with
the same frozen video VAE used for RGB to obtain the clean mask latent
\(m_0 = E_{\mathrm{VAE}}(M)\). The mask branch predicts its velocity
from the noisy RGB latent \(z_t\) and the target RGB features described
above. We use the same noise \(\epsilon\) that forms \(z_t\) to
construct the mask velocity target \(\epsilon-m_0\).

Let \(v_m^+\) and \(v_m^-\) denote the mask velocities from the
positive- and negative-text forward passes, respectively. We use
the same inverse-CFG rule and scheduler weight \(w(t)\) as in
Equation~\eqref{eq:rgb-flow-matching}.
\begin{equation}
    \begin{aligned}
        \bar{v}_m
        &= \frac{v_m^+ + (g-1)\operatorname{sg}(v_m^-)}{g},
        \qquad g=4, \\
        \mathcal{L}_m
        &= \mathbb{E}_{t,\epsilon}\!\left[
            w(t)\left\lVert
                \bar{v}_m - (\epsilon-m_0)
            \right\rVert_2^2
        \right].
    \end{aligned}
    \label{eq:mask-flow-matching}
\end{equation}

\paragraph{Mask warm-up and joint fine-tuning.}
Let \(\theta_r\) and \(\theta_m\) denote the RGB and mask LoRA
parameters, respectively, with \(\theta_r\) corresponding to
\(\theta\) in Section~\ref{sec:source-conditioning}. During mask
warm-up, we keep the trained RGB adapter \(\theta_r^{(0)}\) fixed
and update the mask adapter using \(\mathcal{L}_m\). We then
initialize joint fine-tuning with the trained RGB adapter and the
warmed-up mask adapter, and update both using the combined loss.
\begin{equation}
    \begin{aligned}
        \text{Mask warm-up}\quad
        &\min_{\theta_m}\;
            \mathcal{L}_m(\theta_r^{(0)},\theta_m), \\
        \text{Joint fine-tuning}\quad
        &\min_{\theta_r,\theta_m}\;
            \mathcal{L}_{\mathrm{joint}}(\theta_r,\theta_m), \\
        \mathcal{L}_{\mathrm{joint}}(\theta_r,\theta_m)
        &= \mathcal{L}_{\mathrm{RGB}}(\theta_r)
            + 0.1\,\mathcal{L}_m(\theta_r,\theta_m).
    \end{aligned}
    \label{eq:rgb-mask-training}
\end{equation}

During joint fine-tuning, the target RGB keys and values from the
positive-text forward pass remain in the computation graph. The mask
loss therefore sends gradients to the RGB adapter through the same
attention connection that provides RGB features to the mask branch.
The two adapters receive the following gradients:
\begin{equation}
    \begin{aligned}
        \nabla_{\theta_r}\mathcal{L}_{\mathrm{joint}}
        &= \nabla_{\theta_r}\mathcal{L}_{\mathrm{RGB}}
            + 0.1\,\nabla_{\theta_r}\mathcal{L}_m, \\
        \nabla_{\theta_m}\mathcal{L}_{\mathrm{joint}}
        &= 0.1\,\nabla_{\theta_m}\mathcal{L}_m.
    \end{aligned}
    \label{eq:rgb-mask-gradients}
\end{equation}

\subsection{ACE Features and Ground-Truth Keypoint Supervision}
\label{sec:physical-supervision}
\label{sec:ego-hand-supervision}
\label{sec:model-training}

Downstream operation pipelines often consume 2D hand keypoints extracted
from video, so generated hands should support consistent joint
localization and motion recovery. After comparing candidate methods,
we found ACE-Ego-Hand~\cite{liu2026ace} to be a suitable approach for
directly obtaining 2D hand keypoints from a single egocentric RGB video.
We also use its intermediate hand features as a representation-level
supervision signal.

Our hand supervision combines real annotated data with ACE-derived
pseudo-labels and features. Ground-truth annotations anchor joint
localization and refine the hand reader, while ACE pseudo-labels extend
supervision to recorded EGO clips without manual keypoint annotations.
Intermediate-feature alignment complements coordinate-level targets by
transferring the extractor's hand representation to the generator.

The experiments in Section~\ref{sec:hand-generation-eval} report stronger
ACE-based keypoint agreement and maintained or improved RGB fidelity
for the evaluated hand-supervised training recipes. These results
support the effectiveness of combining GT-refined hand readouts with
ACE pseudo-labels and intermediate-feature supervision.

\paragraph{Features and joint targets}
For recorded EGO video $Y$, frozen ACE-Ego-Hand supplies intermediate
features $F^*$ and decoded pseudo-labels: joint heatmaps $H^*$,
2D coordinates $J^*$, and hand presence $s^*$. Features capture hand
appearance and structure, while the decoded pseudo-labels explicitly
constrain localization and detection.
For predicted features $\widehat F$ and their ACE decoding, we use
\begin{equation}
\begin{aligned}
\mathcal L_{\mathrm{hand}}
={}&\mathcal L_{\mathrm{feat}}(\widehat F,F^*)
+\lambda_H\mathcal L_{\mathrm{KL}}(\widehat H,H^*)\\
&+\lambda_J\mathcal L_{\mathrm{Huber}}(\widehat J,J^*)
+\lambda_s\mathcal L_{\mathrm{BCE}}(\widehat s,s^*).
\end{aligned}
\label{eq:pseudo-hand-loss}
\end{equation}
Feature matching emphasizes hand regions, heatmaps retain spatial
uncertainty, coordinates anchor joint locations, and presence targets
penalize missing or spurious hands.

A projector $P_\psi$ maps H3 EGO latents to ACE features, which the fixed
ACE decoder $D$ converts to hand observations:
\begin{equation}
\widetilde F=P_\psi(z_0),\qquad
(\widetilde H,\widetilde J,\widetilde s)=D(\widetilde F).
\label{eq:hand-reader}
\end{equation}
The projector is aligned to ACE features and refined using hand
annotations from HOT3D~\cite{banerjee2025hot3d},
ARCTIC~\cite{fan2023arctic}, and
Re:InterHand~\cite{moon2023reinterhand}.
Calibrated annotations supply 2D targets and heatmaps:
\begin{equation}
\mathcal L_{\mathrm{GT}}=
\mathcal L_{\mathrm{KL}}(\widetilde H,H^{\mathrm{GT}})
+\beta_J\mathcal L_{\mathrm{Huber}}(\widetilde J,J^{\mathrm{GT}}).
\label{eq:gt-hand-loss}
\end{equation}
ACE-derived pseudo-targets and dataset GT are distinct supervision
sources. Annotated EGO-only clips refine the reader; paired EGO--EXO
clips train the view-conversion generator.

\paragraph{Supervision of the predicted video}
\label{sec:aceaux-training}
\textbf{ACEaux} applies the prepared reader to the predicted clean latent
\begin{equation}
\widehat z_0=z_t-\sigma_t\bar v_\theta(z_t,t;\mathcal C),\qquad
\widehat F=P_{\psi^*}(\widehat z_0).
\label{eq:predicted-clean}
\end{equation}
RGB LoRA is trained with
\begin{equation}
\mathcal L_{\mathrm{ACEaux}}=
\mathcal L_{\mathrm{RGB}}+\lambda_h\mathcal L_{\mathrm{hand}}.
\label{eq:rgb-hand-objective}
\end{equation}
The reader and ACE decoder are fixed, while gradients pass through them
to RGB LoRA. This supervises the structure and motion of the video
currently predicted by the generator.

\paragraph{Supervision of intermediate features}
\label{sec:multilayer-training}
\textbf{MultiLayer} reads hands from several generator depths, combining
spatial evidence with broader action context. Normalized EGO features
are projected, fused with learned weights, and aligned to the ACE grid:
\begin{equation}
\widehat F=A_\phi\!\left(\sum_{\ell\in\mathcal S}
\alpha_\ell W_\ell\operatorname{LN}(h_\ell)\right),
\qquad \sum_\ell\alpha_\ell=1.
\label{eq:multilayer-hand-fusion}
\end{equation}
We first learn the readout with RGB fixed, then jointly adapt it and
RGB LoRA using RGB and hand losses. The hand gradient acts directly on
the selected EGO features. Both routes leave the inference inputs
unchanged, and their hand readers can be removed for RGB generation.

\section{Experiments}
\label{sec:evaluation}

We evaluate RGB fidelity together with hand and object observations
recoverable from generated video. The experiments compare EgoX baselines and examine semantic
joint training, hand-supervision strategies, and camera conditioning.

\subsection{Evaluation Protocol}
Following the seen-/unseen-setting evaluation protocol of EgoX~\cite{kang2025egox},
we assess generalization using our own training and validation splits.
Seen and unseen refer to physical scene settings represented in or held
out from our task-adaptation training split, respectively. The fixed
development-validation panel contains 546 clips spanning 86 captures:
475 seen-setting clips and 71 unseen-setting clips.
All models in each comparison are scored on the same clips. All RGB scores use
$448\times448$ frames at 30\,fps without temporal resampling.
PSNR, SSIM, LPIPS, and CLIP-I average frames within each clip and
then weight clips equally; FVD uses the same I3D implementation
on each complete subset. Our auxiliary-supervision comparisons use
the same frozen EXO/text conditions, 141 output frames, 50 sampling
steps, inference CFG 1, and seed 42.

Table~\ref{tab:egox-rgb-semantic} compares VBot Exo2Ego Policy with
released EgoX~\cite{kang2025egox} and an in-house H3 adaptation,
EgoX/H3. Both EgoX baselines use target-view geometric priors;
VBot Exo2Ego Policy uses EXO video and text without a geometric
prior at inference. EgoX takes 49 EXO frames, while EgoX/H3 and our
model take 141; the shorter evaluation of our model uses the first
49 frames of the same generated video. Our model is the Joint v1
checkpoint after 3,000 updates, and EgoX/H3 is evaluated at step
3,000. The comparison therefore evaluates the complete systems
under their respective conditioning and training recipes.

\begin{table}[H]
\centering
\caption{\textbf{VBot Exo2Ego Policy compared with EgoX for first-person video generation.}
EgoX is the released model; EgoX/H3 is our H3-based adaptation.
VBot Exo2Ego Policy achieves better reconstruction, perceptual similarity,
and video-distribution quality across both clip lengths and all subsets.}
\label{tab:egox-rgb-semantic}
\fontsize{8.3}{10.1}\selectfont
\setlength{\tabcolsep}{4pt}
\renewcommand{\arraystretch}{1.08}
\begin{tabular}{@{}llrrrrr@{}}
\toprule
\textbf{Subset} & \textbf{Model} & \textbf{PSNR}$\uparrow$ & \textbf{SSIM}$\uparrow$ & \textbf{LPIPS}$\downarrow$ & \textbf{CLIP-I}$\uparrow$ & \textbf{FVD}$\downarrow$ \\
\midrule
\multicolumn{7}{@{}l}{\textcolor{deepnavy}{\textbf{49-frame clips}}} \\
\addlinespace[3pt]
All (546) & EgoX & 12.630 & 0.4087 & 0.5942 & 0.8427 & 306.90 \\
 & VBot Exo2Ego Policy & \textbf{15.856} & \textbf{0.4894} & \textbf{0.4955} & \textbf{0.8708} & \textbf{191.55} \\
\addlinespace[2pt]
Seen (475) & EgoX & 12.735 & 0.4186 & 0.5903 & 0.8440 & 322.06 \\
 & VBot Exo2Ego Policy & \textbf{15.897} & \textbf{0.4943} & \textbf{0.4923} & \textbf{0.8724} & \textbf{199.79} \\
\addlinespace[2pt]
Unseen (71) & EgoX & 11.927 & 0.3422 & 0.6209 & 0.8338 & 1146.71 \\
 & VBot Exo2Ego Policy & \textbf{15.584} & \textbf{0.4568} & \textbf{0.5174} & \textbf{0.8600} & \textbf{908.01} \\
\midrule
\multicolumn{7}{@{}l}{\textcolor{deepnavy}{\textbf{141-frame clips}}} \\
\addlinespace[3pt]
All (545) & EgoX/H3 & 12.579 & 0.3832 & 0.5971 & 0.8373 & 284.49 \\
 & VBot Exo2Ego Policy & \textbf{15.826} & \textbf{0.4891} & \textbf{0.4954} & \textbf{0.8714} & \textbf{171.21} \\
\addlinespace[2pt]
Seen (474) & EgoX/H3 & 12.598 & 0.3885 & 0.5973 & 0.8378 & 297.36 \\
 & VBot Exo2Ego Policy & \textbf{15.871} & \textbf{0.4942} & \textbf{0.4922} & \textbf{0.8729} & \textbf{170.46} \\
\addlinespace[2pt]
Unseen (71) & EgoX/H3 & 12.449 & 0.3477 & 0.5955 & 0.8338 & 974.04 \\
 & VBot Exo2Ego Policy & \textbf{15.525} & \textbf{0.4551} & \textbf{0.5168} & \textbf{0.8613} & \textbf{782.86} \\
\bottomrule
\end{tabular}
\vspace{4pt}

\parbox{\linewidth}{\footnotesize Each pair of rows uses the same clips and ground truth.
The 141-frame group excludes one EgoX/H3 prior failure from both models;
the following auxiliary-supervision tables retain all 546 clips.}
\end{table}

VBot Exo2Ego Policy improves all five metrics in every subset.
Relative to released EgoX, it gains 3.23\,dB in PSNR and reduces FVD
by 37.6\%; relative to EgoX/H3, the gains are 3.25\,dB and 39.8\%.

Native ACE independently processes generated and real EGO RGB under
the same per-clip calibration and anatomical left/right identities.
PCK thresholds are 5\% and 10\% of image width; missing predictions
count as failures. EPE and joint-motion error are reported with
coverage. SAM3 object masks are evaluated against official sparse
annotations (91 clips, 170 object frames) and a real-RGB SAM reference
(247 positive clips, 1,449 frames). These readouts evaluate generated
RGB, separately from auxiliary-head accuracy.

The auxiliary-supervision tables compare training recipes with
different optimization budgets. ACE joints are pseudo-references,
while dataset GT trains
and validates the hand reader. This development panel has been used
for checkpoint selection. Full scoring and training details are in
Appendix~\ref{app:training-details}.

\subsection{Semantic Supervision}
\label{sec:semantic-results}
Independent hand and object readouts show improved localization with
semantic joint training (Table~\ref{tab:semantic-hoi-generation}).
PCK$_{.05}$ increases by 5.019 percentage points and official object IoU
by 2.299 points.
Joint coverage changes from 92.433\% to 90.805\%; the gain concerns
localization rather than uniform improvement across every hand metric.

\begin{table}[H]
\centering
\caption{\textbf{Hand and object readouts from generated RGB (\%).}}
\label{tab:semantic-hoi-generation}
\fontsize{8.3}{10.1}\selectfont
\setlength{\tabcolsep}{5pt}
\begin{tabular}{@{}lrrrrr@{}}
\toprule
\textbf{Model} & \textbf{PCK}$_{.05}\uparrow$ & \textbf{PCK}$_{.10}\uparrow$ & \textbf{Coverage}$\uparrow$ & \textbf{Official IoU}$\uparrow$ & \textbf{SAM IoU}$\uparrow$ \\
\midrule
RGB-only & 11.519 & 39.773 & \textbf{92.433} & 5.471 & 7.015 \\
RGB + Semantic & \textbf{16.538} & \textbf{48.346} & 90.805 & \textbf{7.770} & \textbf{8.584} \\
\bottomrule
\end{tabular}
\end{table}

\subsection{Hand Supervision}
\label{sec:hand-generation-eval}

\begin{table}[H]
\centering
\caption{\textbf{RGB quality under different hand-supervision strategies.}}
\label{tab:hand-rgb-quality}
\small
\setlength{\tabcolsep}{4pt}
\renewcommand{\arraystretch}{1.13}
\begin{tabularx}{\linewidth}{@{}Xrrrrr@{}}
\toprule
\textbf{Model} & PSNR $\uparrow$ & SSIM $\uparrow$ & LPIPS $\downarrow$ & CLIP-I $\uparrow$ & FVD $\downarrow$ \\
\midrule
\multicolumn{6}{@{}l}{\textcolor{deepnavy}{\textbf{Seen: 475 clips, 79 captures}}} \\
RGB-only & 15.363 & 0.4823 & 0.5173 & 0.8657 & 202.06 \\
ACEaux & 15.943 & \textbf{0.5015} & 0.4930 & 0.8668 & 180.95 \\
MultiLayer & \textbf{15.977} & 0.5003 & \textbf{0.4925} & \textbf{0.8680} & \textbf{171.73} \\
\addlinespace[3pt]
\multicolumn{6}{@{}l}{\textcolor{deepnavy}{\textbf{Unseen: 71 clips, 7 captures}}} \\
RGB-only & 14.939 & 0.4442 & 0.5429 & 0.8546 & 845.76 \\
ACEaux & 15.529 & \textbf{0.4632} & \textbf{0.5192} & 0.8525 & 840.49 \\
MultiLayer & \textbf{15.647} & 0.4623 & 0.5198 & \textbf{0.8590} & \textbf{805.11} \\
\addlinespace[3pt]
\multicolumn{6}{@{}l}{\textcolor{deepnavy}{\textbf{Total: 546 clips, 86 captures}}} \\
RGB-only & 15.308 & 0.4774 & 0.5206 & 0.8643 & 204.38 \\
ACEaux & 15.890 & \textbf{0.4965} & 0.4964 & 0.8649 & 187.37 \\
MultiLayer & \textbf{15.934} & 0.4954 & \textbf{0.4961} & \textbf{0.8668} & \textbf{173.28} \\
\bottomrule
\end{tabularx}
\end{table}

\begin{table}[H]
\centering
\caption{\textbf{Hand structure and motion extracted from generated RGB.}}
\label{tab:hand-generated-metrics}
\small
\setlength{\tabcolsep}{4pt}
\renewcommand{\arraystretch}{1.13}
\begin{tabularx}{\linewidth}{@{}Xrrrrrr@{}}
\toprule
\textbf{Model} & \shortstack{PCK@5\%\\$\uparrow$} & \shortstack{PCK@10\%\\$\uparrow$} & \shortstack{EPE\\(px) $\downarrow$} & \shortstack{Joint cov.\\(\%) $\uparrow$} & \shortstack{Motion err.\\(px/fr) $\downarrow$} & \shortstack{Motion cov.\\(\%) $\uparrow$} \\
\midrule
\multicolumn{7}{@{}l}{\textcolor{deepnavy}{\textbf{Seen: 475 clips, 79 captures}}} \\
RGB-only & 12.47 & 41.87 & 55.86 & 91.72 & 4.447 & 91.37 \\
ACEaux & \textbf{21.44} & \textbf{56.51} & \textbf{46.51} & \textbf{94.62} & 4.214 & \textbf{94.40} \\
MultiLayer & 18.26 & 51.14 & 47.36 & 88.71 & \textbf{4.206} & 88.31 \\
\addlinespace[3pt]
\multicolumn{7}{@{}l}{\textcolor{deepnavy}{\textbf{Unseen: 71 clips, 7 captures}}} \\
RGB-only & 5.24 & 25.92 & 65.13 & 97.12 & 3.452 & 97.01 \\
ACEaux & \textbf{10.64} & \textbf{42.45} & \textbf{53.41} & \textbf{98.92} & \textbf{3.184} & \textbf{98.87} \\
MultiLayer & 10.33 & 39.54 & 55.68 & 96.36 & 3.475 & 96.27 \\
\addlinespace[3pt]
\multicolumn{7}{@{}l}{\textcolor{deepnavy}{\textbf{Total: 546 clips, 86 captures}}} \\
RGB-only & 11.52 & 39.77 & 57.09 & 92.43 & 4.316 & 92.12 \\
ACEaux & \textbf{20.02} & \textbf{54.66} & \textbf{47.42} & \textbf{95.18} & \textbf{4.078} & \textbf{94.99} \\
MultiLayer & 17.21 & 49.61 & 48.46 & 89.72 & 4.108 & 89.36 \\
\bottomrule
\end{tabularx}
\end{table}

Native ACE is run independently on generated RGB and recorded EGO to
measure recoverable hand structure (Table~\ref{tab:hand-generated-metrics}).
ACEaux gives the strongest extraction agreement: PCK@5\% increases
from 11.52\% to 20.02\%, PCK@10\% from 39.77\% to 54.66\%, and
joint coverage from 92.43\% to 95.18\%. EPE decreases from 57.09 to
47.42 pixels. MultiLayer gives the strongest aggregate image metrics,
with higher PCK but lower hand coverage than the RGB-only reference.

\subsection{Camera-Conditioning Optimization}

At native 256P, stochastic ray conditioning followed by ray-free
continuation yields lower logged flow-MSE than the historical RGB
reference (Table~\ref{tab:ray-training-loss}). During updates
3001--5000, mean loss is 0.12557 versus 0.13638, a 7.93\% reduction.
These are optimization measurements under the recorded recipes;
the loss curves and settings appear in Appendix~\ref{app:training-details}.

\begin{table}[H]
\centering
\caption{\textbf{Weighted training flow-MSE at native 256P.} Values average the original per-update losses.}
\label{tab:ray-training-loss}
\small
\setlength{\tabcolsep}{7pt}
\renewcommand{\arraystretch}{1.13}
\begin{tabularx}{\linewidth}{@{}Xrrr@{}}
\toprule
\textbf{Updates} & \textbf{Historical RGB} & \textbf{Stochastic rays} & \textbf{Relative change} \\
\midrule
1--3000 & 0.14070 & 0.13051 & -7.24\% \\
3001--5000 & 0.13638 & 0.12557 & -7.93\% \\
4501--5000 & 0.13712 & 0.12383 & -9.69\% \\
4901--5000 & 0.14002 & 0.12436 & -11.19\% \\
1--5000 & 0.13897 & 0.12853 & -7.51\% \\
\bottomrule
\end{tabularx}
\end{table}

\subsection{Qualitative Results}

Figure~\ref{fig:auxiliary-outputs} shows interaction masks and hand
observations from generated EGO. The hand panels compare direct H3
readout with independent ACE extraction on the same generated frames.
Internet examples retain horizontal EXO/EGO pairs
(Figure~\ref{fig:internet-blocking}). Multi-keyframe semantic results
and GT/ACE/projector comparisons on recorded EGO are collected in
Appendix~\ref{app:qualitative-results}.

\begingroup
\setlength{\intextsep}{5pt}
\captionsetup{skip=4pt}
\begin{figure}[H]
\centering
\includegraphics[width=0.97\linewidth]{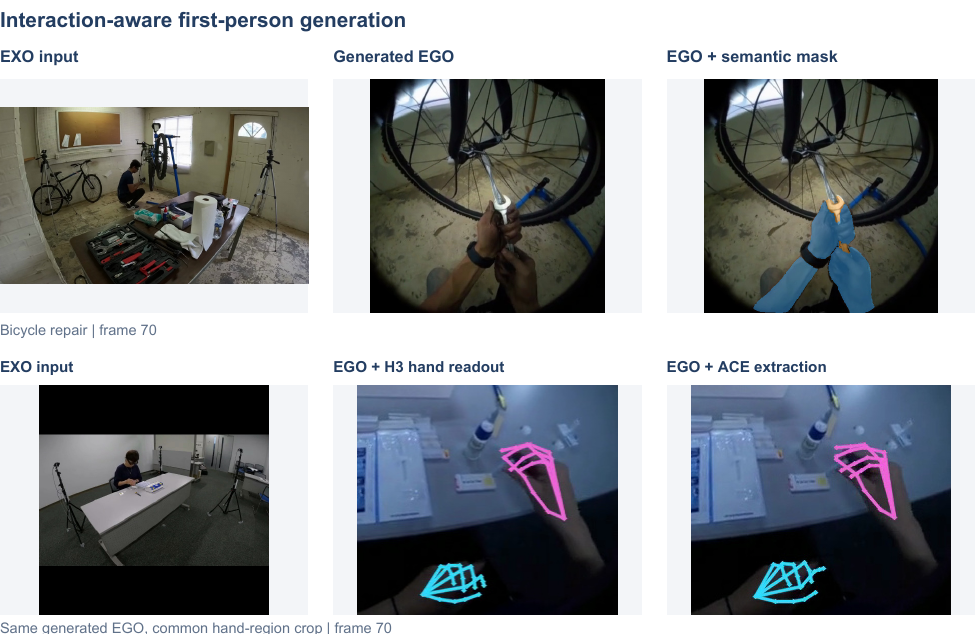}
\caption{\textbf{Semantic masks and hand readouts from generated EGO.}}
\label{fig:auxiliary-outputs}
\end{figure}

\begin{figure}[H]
\centering
\includegraphics[width=0.97\linewidth]{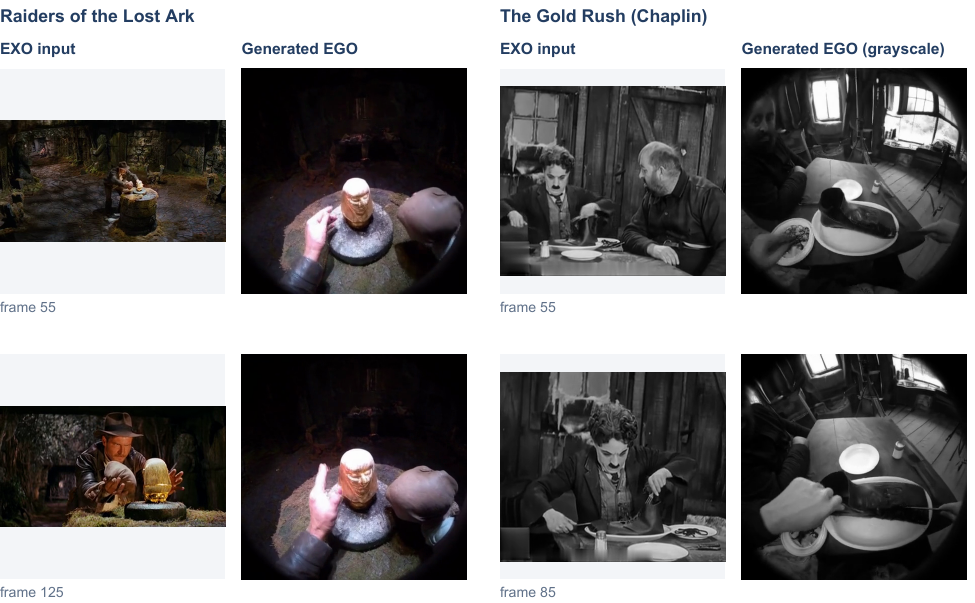}
\caption{\textbf{Internet EXO-to-EGO examples.}}
\label{fig:internet-blocking}
\label{fig:internet-handover}
\end{figure}
\endgroup

\section{Conclusion and Limitations}
\label{sec:conclusion}

VBot Exo2Ego Policy connects lightweight data collection with first-person
video generation. Precise paired capture supplies camera and appearance
correspondence, while single-view crowdsourcing expands real-world
coverage. Stochastic camera conditioning supports viewpoint learning;
semantic HOI and ACE/GT hand objectives connect RGB generation to
interaction regions and extractable joint structure. The evaluated
auxiliary-training recipes improve these interaction readouts alongside
reported RGB fidelity. Source reasoning, actor selection, and multi-shot
training extend the same model interface to diverse real videos.

\paragraph{Limitations}\label{sec:limitations}

EXO footage can hide hand articulation, object surfaces, and head motion,
leaving several plausible EGO observations. Auxiliary supervision
inherits uncertainty from its labels and extractors, and performance
may vary across devices, tasks, and viewpoints. Synchronized camera
switches cover a continuous event; arbitrary time skips and actor
changes require broader edited-video training.

\par
\clearpage
\appendix
\section{Training and Evaluation Details}
\label{app:training-details}
\label{sec:deployment}
\label{sec:reproducibility}
\label{app:evidence-map}

\subsection{Shared Flow Convention}
The effective velocity uses the base model's inverse-CFG convention,
\[
\bar v=\frac{v^++(g-1)\operatorname{sg}(v^-)}{g},\qquad g=4,
\]
with positive and negative text predictions sharing the noisy target
and source references. The negative prediction is stop-gradient.
Inference CFG is 1. Each training recipe retains its scheduler weight
and latent/time alignment.

\subsection{Geometric Conditioning}
The reported ray experiment uses 33,056 clips at native 256P, batch
size 8, learning rate $10^{-4}$, and AdamW weight decay 0.01.
The initial 3,000-update schedule is followed by 2,000 ray-free updates.
The historical RGB reference uses a different data pool and learning-rate
schedule, so its loss comparison is descriptive. A separate cosine
schedule produced a transient late loss burst; the reported linear
schedule and ray-free continuation remained finite.

\subsection{Semantic Joint Training}
Starting from pretrained MiniMax-H3 Ref2VA, the preceding paired RGB
task-adaptive stage configures 210,261 cache entries from Lance v3/v4
and v5 (35,927 and 174,334, respectively). These are configured paired
entries, not distinct EGO targets or a claim that every entry has been
visited by the selected checkpoint. Targets have 141 frames at
$448\times448$. The base remains frozen while rank-32 RGB LoRA is
trained at learning rate $2\times10^{-5}$ with global batch size 8.
The RGB-only reference is selected at update 14,000 of this
continuation; this local counter is not the cumulative update count
of all preceding training.

The semantic subset contains 13,188 distinct EGO temporal windows
and 49,012 candidate EXO pairings. Each epoch samples EGO windows and
chooses one available EXO view per window. Semantic videos retain
the EGO RGB time and pixel coordinates. Their fixed palette uses black
for background, orange for manipulated objects, and blue for wearer
hands/arms/sleeves. They are encoded with the same frozen native VAE.

Mask warm-up holds RGB LoRA fixed and trains rank-32 mask LoRA for
4,000 updates at learning rate $10^{-4}$, optimizing only the mask
velocity term with unit coefficient. Joint fine-tuning starts from
the trained RGB-only adapter and the warmed-up mask adapter. Both
adapters are updated with a fresh AdamW optimizer, 25 updates of
linear learning-rate warm-up followed by a constant learning rate
of $10^{-5}$, global batch size 8, zero weight decay, and gradient-norm
clipping at 1. The base and VAE remain frozen. Both branch predictions
use the shared inverse-CFG convention ($g=4$), with the negative
forward pass evaluated without gradients. The mask coefficient is
0.1; clean-latent reconstruction and audio losses have zero weight.
The quantitative checkpoint is Joint v1 after 3,000 joint updates;
mask warm-up and joint training have separate counters.

Hand and image statistics use equal clip weights. Hand evaluation
has 539 clips with eligible ACE reference points. Manipulated-object
IoU is evaluated separately on official annotations and the real-RGB
SAM pseudo-reference, with positive-reference failures retained in
the denominator. Paired capture-cluster bootstrap uses 10,000
resamples. The PCK$_{.05}$ and official-IoU gain intervals are
$[3.758,6.096]$ and $[1.027,3.592]$ percentage points, respectively.
These 95\% intervals condition on the selected training run and
inference seed; they do not estimate training-seed variation.
Official unseen-object support is limited to 10 clips from four
captures. Continuing the same joint run to 4,900 updates gives FVD
216.336, illustrating checkpoint sensitivity rather than a guarantee
of better generation with longer training.

The qualitative figures use a separate 5,020-update joint checkpoint,
50 sampling steps, and seed 85, including one Echo EXO-only source
and two Ego-Exo4D training clips. They are not examples from the
Joint v1/3000 quantitative panel. The RGB-only scores in the
auxiliary-supervision tables use the measured 14,000-update reference.

\subsection{Hand Supervision}
Both hand-supervised variants adapt RGB for one epoch on 45,524 paired
clips. ACEaux gradually introduces hand loss to weight 0.03.
MultiLayer fuses EGO features from blocks $\{6,12,18,25\}$, prepares the
readout for three epochs, then jointly adapts it with RGB for one epoch.
The shared hand-loss weights are $\lambda_H=0.1$,
$\lambda_J=0.5$, and $\lambda_s=0.01$.

The hand table's RGB-only continuation differs in budget and
learning-rate recipe. Against the shared initial generator, ACEaux
and MultiLayer improve PSNR by 0.111 and 0.155\,dB and reduce LPIPS by
0.0060 and 0.0064. Relative to the listed RGB-only reference, ACEaux's
unseen CLIP-I is slightly lower; MultiLayer's joint coverage is lower.
Motion error measures agreement with reference displacement.
Reader illustrations compare annotation-projected GT, native ACE
(K-given), and the H3 projector on matched crops with presence threshold
0.5.

The generated-EGO hand illustration in Figure~\ref{fig:auxiliary-outputs}
uses the 5,020-update RGB/mask model with 50 sampling steps and seed 42.
H3 and native ACE read the same generated sequence; both displayed
EGO panels use the same hand-region crop. Their agreement is a
readout comparison rather than an error against annotated GT.

\subsection{Scoring and Comparison Notes}
ACE readouts use the original 448-pixel fisheye canvas, with PCK
thresholds of 22.4 and 44.8 pixels. There are 539 clips with eligible
real-RGB reference joints. Positive-reference object frames with missing
predictions receive zero overlap; empty references do not count as
successful negatives. Official and SAM-reference object IoU use
different annotated subsets.

The semantic motion-residual interval includes zero. Official
unseen-object support covers 10 clips from four captures and disagrees
in direction with the SAM-reference readout, so no unseen-object gain
is established.

\subsection{Camera-Conditioning Loss Curves}
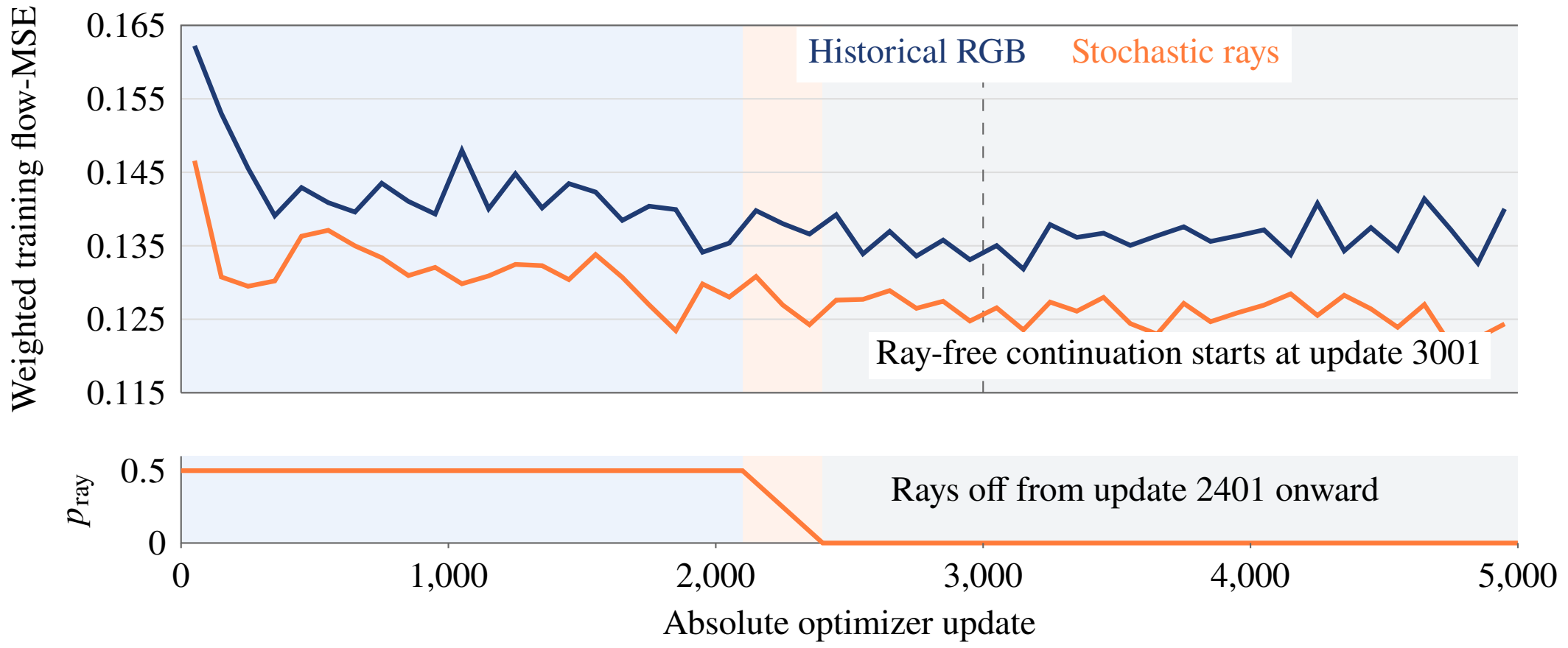
\begin{figure}[H]
\centering
\begin{tikzpicture}[x=12.0cm,y=3.3cm,font=\footnotesize]
\fill[paleblue] (0,0) rectangle (0.42,1);
\fill[paleorange] (0.42,0) rectangle (0.48,1);
\fill[abstractbg] (0.48,0) rectangle (1,1);
\draw[color=rulelinegray!45] (0,0.000)--(1,0.000);
\node[left] at (0,0.000) {0.115};
\draw[color=rulelinegray!45] (0,0.200)--(1,0.200);
\node[left] at (0,0.200) {0.125};
\draw[color=rulelinegray!45] (0,0.400)--(1,0.400);
\node[left] at (0,0.400) {0.135};
\draw[color=rulelinegray!45] (0,0.600)--(1,0.600);
\node[left] at (0,0.600) {0.145};
\draw[color=rulelinegray!45] (0,0.800)--(1,0.800);
\node[left] at (0,0.800) {0.155};
\draw[color=rulelinegray!45] (0,1.000)--(1,1.000);
\node[left] at (0,1.000) {0.165};
\draw[color=softgray] (0,0)--(0,1)--(1,1);
\draw[color=softgray] (0,0)--(1,0);
\draw[dashed,color=softgray] (0.6,0)--(0.6,1);
\draw[very thick,color=deepnavy] plot coordinates {(0.010100,0.944356) (0.030100,0.760082) (0.050100,0.611164) (0.070100,0.481788) (0.090100,0.558466) (0.110100,0.517474) (0.130100,0.491876) (0.150100,0.570310) (0.170100,0.520777) (0.190100,0.486487) (0.210100,0.659567) (0.230100,0.500805) (0.250100,0.596023) (0.270100,0.503037) (0.290100,0.569173) (0.310100,0.546253) (0.330100,0.469140) (0.350100,0.507700) (0.370100,0.498654) (0.390100,0.382495) (0.410100,0.407344) (0.430100,0.495700) (0.450100,0.460243) (0.470100,0.431930) (0.490100,0.484589) (0.510100,0.378085) (0.530100,0.439133) (0.550100,0.372010) (0.570100,0.415504) (0.590100,0.362042) (0.610100,0.400523) (0.630100,0.336898) (0.650100,0.457716) (0.670100,0.422873) (0.690100,0.434331) (0.710100,0.401046) (0.730100,0.427149) (0.750100,0.451688) (0.770100,0.411899) (0.790100,0.426988) (0.810100,0.443351) (0.830100,0.375632) (0.850100,0.515385) (0.870100,0.386138) (0.890100,0.449378) (0.910100,0.387876) (0.930100,0.527539) (0.950100,0.442997) (0.970100,0.352774) (0.990100,0.500466)};
\draw[very thick,color=warmorange] plot coordinates {(0.010100,0.631514) (0.030100,0.315099) (0.050100,0.289776) (0.070100,0.304249) (0.090100,0.426053) (0.110100,0.442118) (0.130100,0.399814) (0.150100,0.367526) (0.170100,0.318930) (0.190100,0.341218) (0.210100,0.296527) (0.230100,0.317991) (0.250100,0.349297) (0.270100,0.345892) (0.290100,0.307915) (0.310100,0.376225) (0.330100,0.314164) (0.350100,0.239427) (0.370100,0.168907) (0.390100,0.296098) (0.410100,0.260369) (0.430100,0.316394) (0.450100,0.238808) (0.470100,0.185078) (0.490100,0.252073) (0.510100,0.254177) (0.530100,0.277995) (0.550100,0.229743) (0.570100,0.248982) (0.590100,0.195288) (0.610100,0.231149) (0.630100,0.171075) (0.650100,0.246699) (0.670100,0.222006) (0.690100,0.259515) (0.710100,0.188246) (0.730100,0.159500) (0.750100,0.243246) (0.770100,0.193290) (0.790100,0.217355) (0.810100,0.238454) (0.830100,0.269144) (0.850100,0.210458) (0.870100,0.265349) (0.890100,0.228351) (0.910100,0.178110) (0.930100,0.240121) (0.950100,0.128310) (0.970100,0.149408) (0.990100,0.187221)};
\node[rotate=90] at (-0.115,0.5) {Weighted training flow-MSE};
\node[anchor=west,fill=white,inner sep=3pt] at (0.46,0.92) {\textcolor{deepnavy}{Historical RGB} \quad \textcolor{warmorange}{Stochastic rays}};
\node[anchor=south east,fill=white,inner sep=2pt] at (0.98,0.035) {Ray-free continuation starts at update 3001};
\begin{scope}[yshift=-1.35cm,y=1.3cm]
\fill[paleblue] (0,0) rectangle (0.42,0.6);
\fill[paleorange] (0.42,0) rectangle (0.48,0.6);
\fill[abstractbg] (0.48,0) rectangle (1,0.6);
\draw[color=softgray] (0,0.6)--(0,0)--(1,0);
\draw[very thick,color=warmorange] (0,0.5)--(0.42,0.5)--(0.48,0)--(1,0);
\node[left] at (0,0.5) {$0.5$};
\node[left] at (0,0) {$0$};
\node[rotate=90] at (-0.075,0.3) {$p_{\mathrm{ray}}$};
\node[anchor=west] at (0.52,0.35) {Rays off from update 2401 onward};
\draw[color=softgray] (0.0,0)--(0.0,-0.035);
\node[below] at (0.0,-0.035) {0};
\draw[color=softgray] (0.2,0)--(0.2,-0.035);
\node[below] at (0.2,-0.035) {1,000};
\draw[color=softgray] (0.4,0)--(0.4,-0.035);
\node[below] at (0.4,-0.035) {2,000};
\draw[color=softgray] (0.6,0)--(0.6,-0.035);
\node[below] at (0.6,-0.035) {3,000};
\draw[color=softgray] (0.8,0)--(0.8,-0.035);
\node[below] at (0.8,-0.035) {4,000};
\draw[color=softgray] (1.0,0)--(1.0,-0.035);
\node[below] at (1.0,-0.035) {5,000};
\node[below] at (0.5,-0.36) {Absolute optimizer update};
\end{scope}
\end{tikzpicture}
\vspace{0.5em}
\caption{\textbf{Training loss and ray availability.} The curves average consecutive 100-update blocks; conditioning probability follows the original schedule during ray-free continuation.}
\label{fig:ray-training-loss}
\end{figure}

\section{Data Alignment and Packaging}
\label{app:alignment-packaging}
\label{sec:synchronization}
\label{sec:model-interface}

Paired capture timestamps are mapped to a common time base. The
nearest EGO frame to each EXO reference is selected within the recorded
20\,ms acceptance window. Camera calibration and image transforms remain
aligned to the decoded frames. Dataset records bind source clips,
view provenance, text, optional camera rays, and auxiliary targets to
their corresponding temporal windows.

The model uses 141-frame clips. Capture timestamps and playback rate
are recorded separately, including the 30\,fps dataset and 24\,fps H3
playback convention. Training and evaluation splits group source
recordings with all derived clips and generated views. Release metadata
identifies the model, preprocessing, annotation, and configuration
versions used to produce each record.

\begin{table}[H]
\centering\small
\caption{\textbf{Inputs and supervision.}}
\label{tab:information-roles}
\begin{tabularx}{\linewidth}{@{}p{0.30\linewidth}XX@{}}
\toprule
\textbf{Information} & \textbf{Training role} & \textbf{Inference role} \\
\midrule
EXO video and source text & Visual and task conditions & Same conditions \\
Actor-reference image & Optional actor selection & Optional actor selection \\
Target-camera rays & Stochastic camera condition & Annealed away \\
Recorded EGO RGB & Video-generation target & Generated output \\
EGO semantic masks & Hand/object region supervision & Optional joint output \\
ACE features and observations & Hand-structure supervision & Training only \\
Annotated hand keypoints & Calibrate hand supervision & Training only \\
\bottomrule
\end{tabularx}
\end{table}

\clearpage
\section{Additional Qualitative Results}
\label{app:qualitative-results}

Joint RGB/mask examples use the separate 5,020-update checkpoint and
seed 85. The recorded-EGO reader comparisons use annotation-projected
GT, native ACE, and the final H3 projector. All EXO/EGO pairs use matched
frame indices.

\begin{figure}[H]
\centering
\includegraphics[page=1,width=\linewidth]{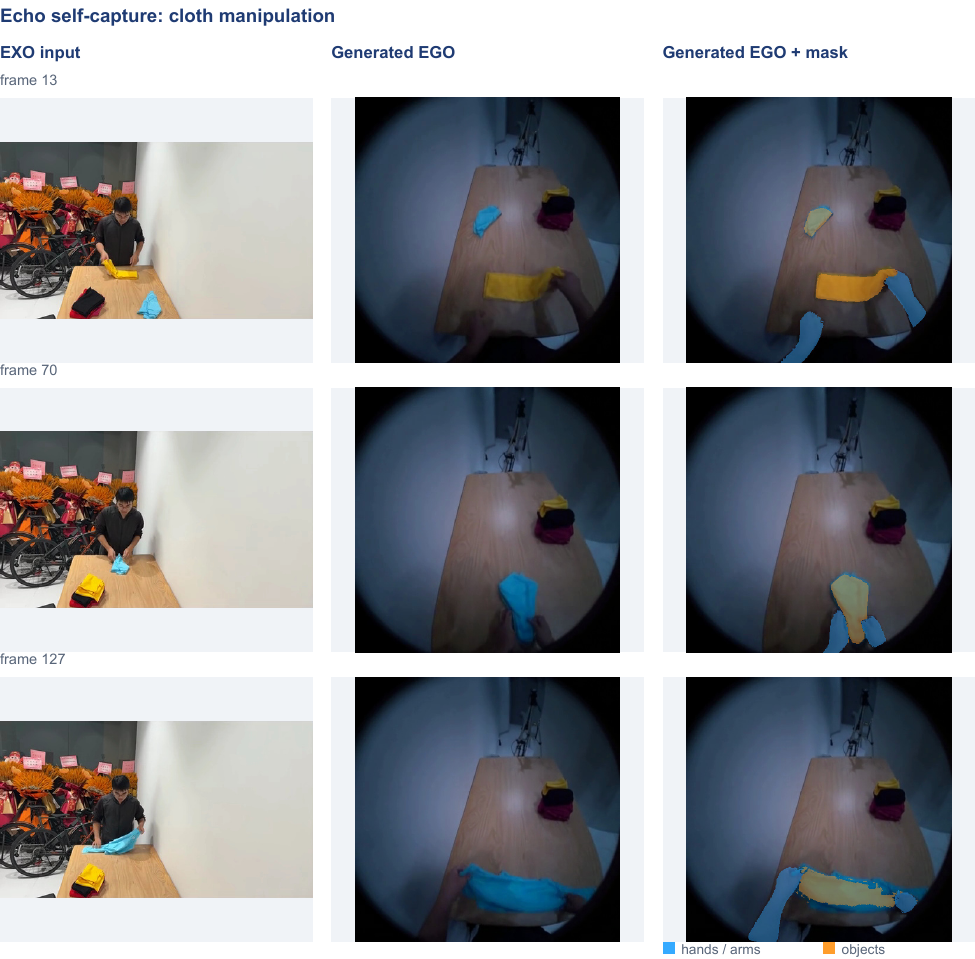}
\caption{\textbf{Joint RGB/mask generation on Echo self-collected data.}}
\label{fig:semantic-echo}
\end{figure}

\begin{figure}[H]
\centering
\includegraphics[page=2,width=\linewidth]{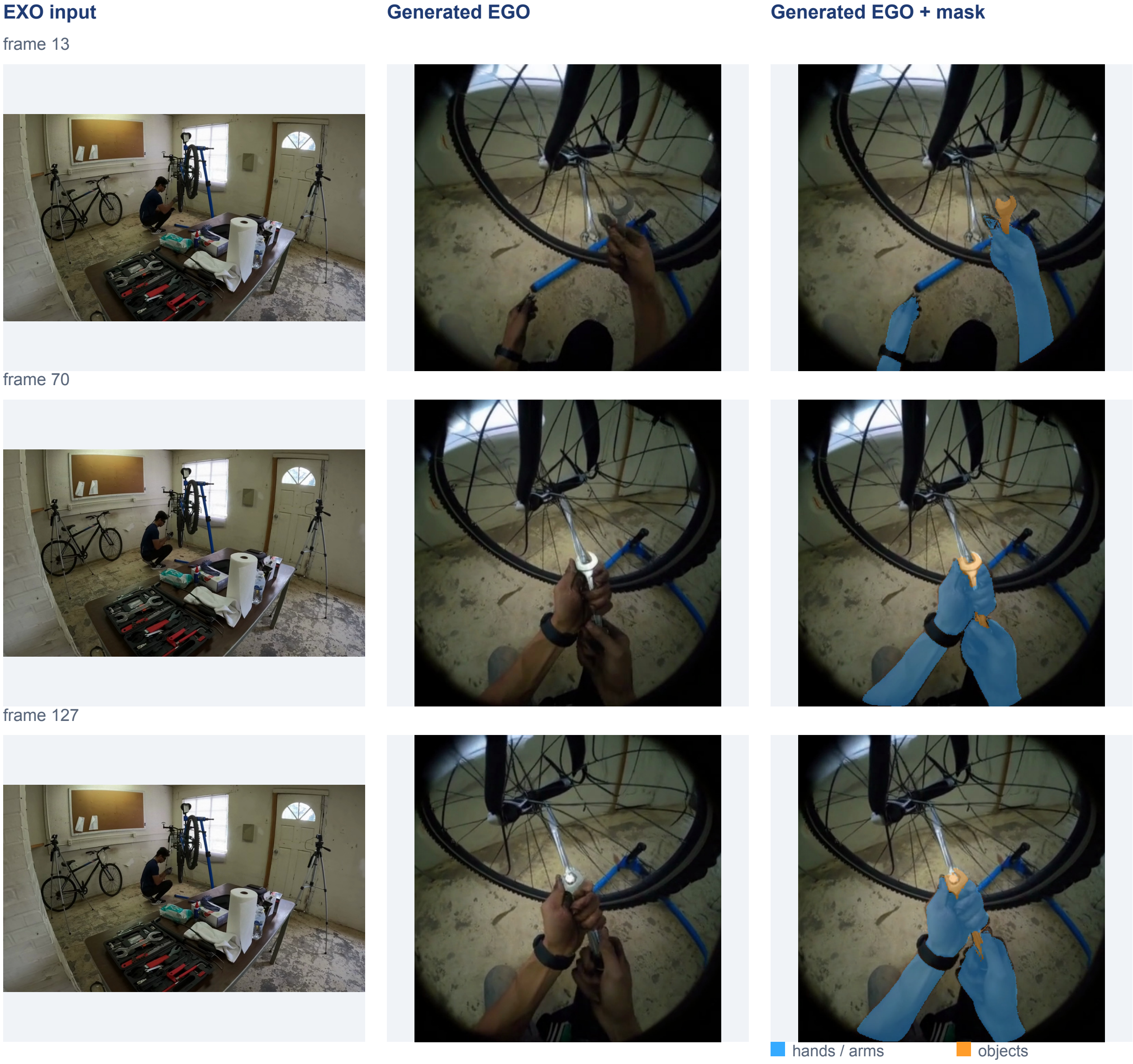}
\caption{\textbf{Joint RGB/mask generation for bicycle repair.}}
\label{fig:semantic-bike}
\end{figure}

\begin{figure}[H]
\centering
\includegraphics[page=3,width=\linewidth]{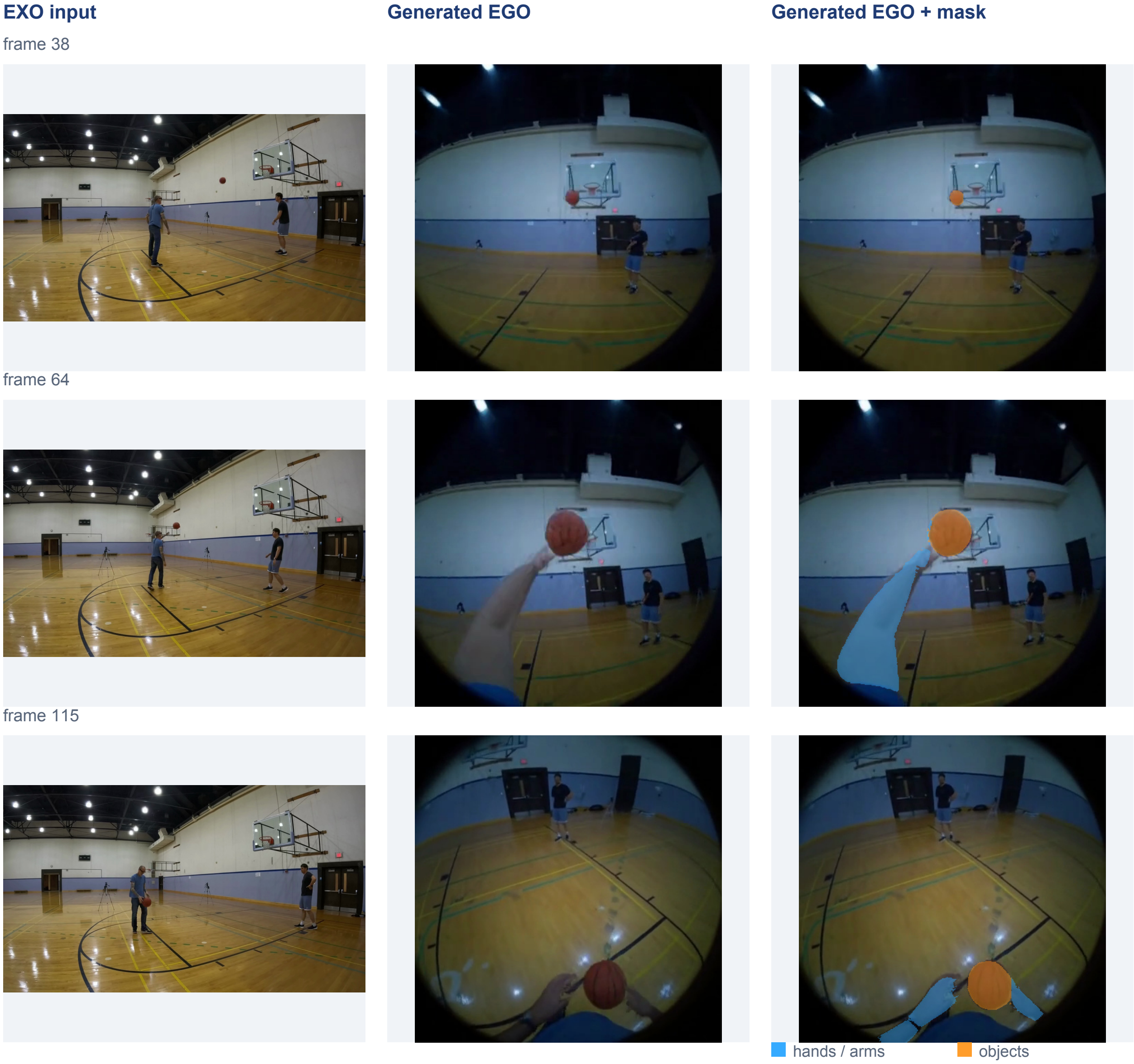}
\caption{\textbf{Joint RGB/mask generation for basketball.}}
\label{fig:semantic-basketball}
\end{figure}

\begin{figure}[H]
\centering
\includegraphics[page=5,width=\linewidth]{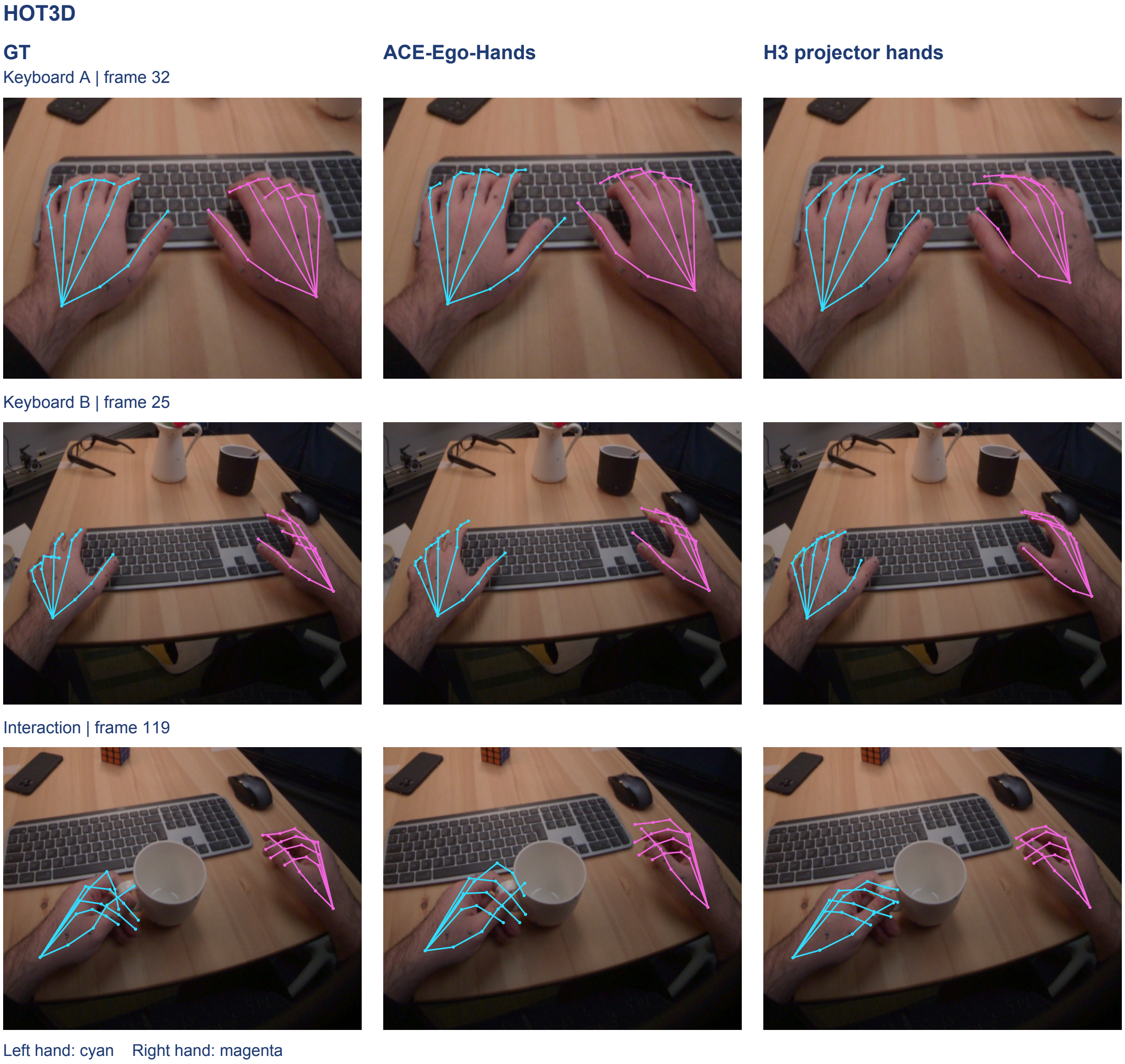}
\caption{\textbf{Hand observations on HOT3D.}}
\label{fig:reader-hot3d}
\end{figure}

\begin{figure}[H]
\centering
\includegraphics[page=4,width=\linewidth]{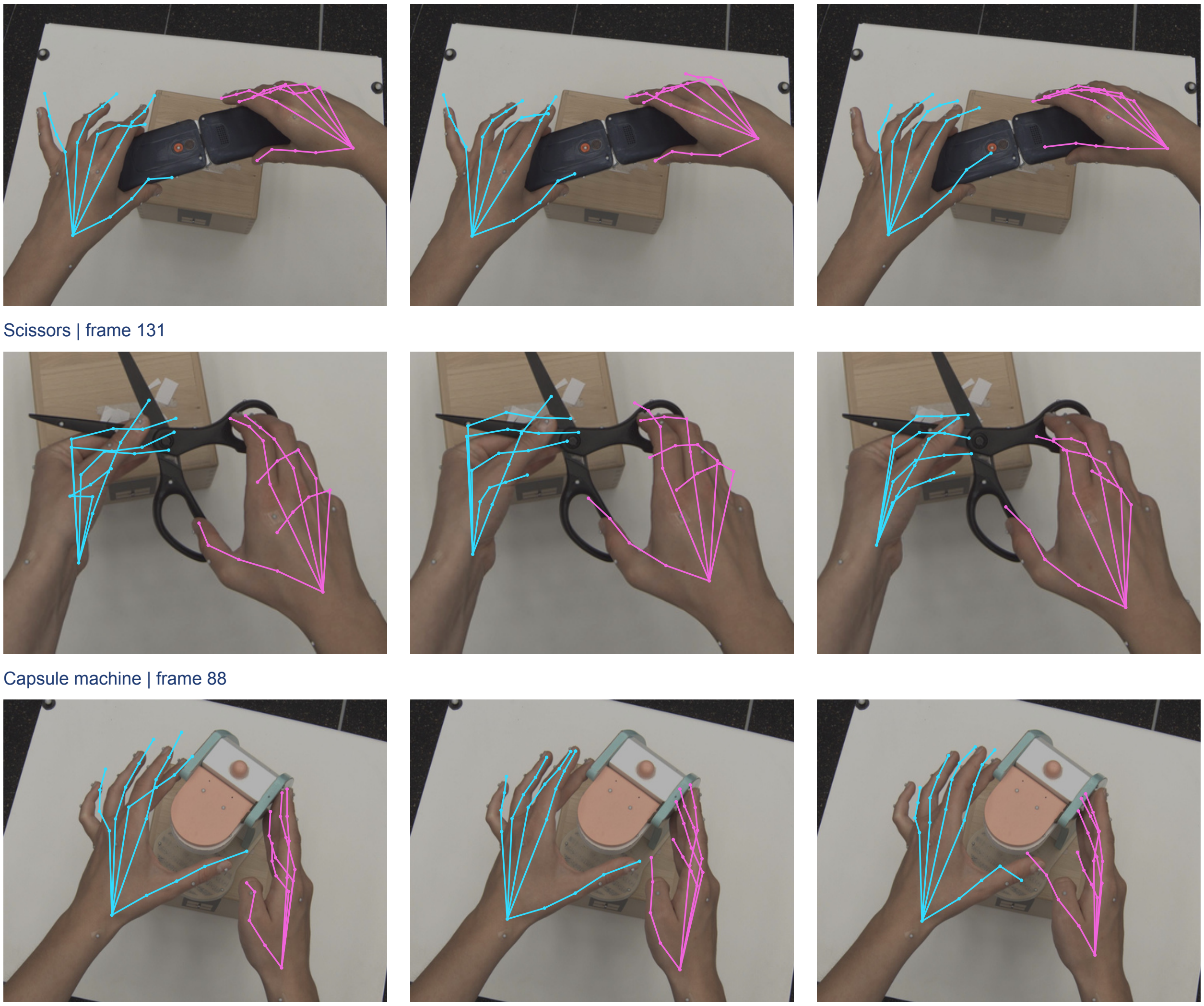}
\caption{\textbf{Hand observations on ARCTIC.}}
\label{fig:reader-arctic}
\end{figure}

\begin{figure}[H]
\centering
\includegraphics[page=7,width=\linewidth]{exo2ego-results-v019.pdf}
\caption{\textbf{EXO-to-EGO generation: \emph{Star Wars: A New Hope}.}}
\label{fig:internet-lightsaber}
\end{figure}

\end{document}